\pdfoutput=1

\documentclass[11pt,a4paper]{article}
\usepackage{pifont}
\usepackage{fourier}
\usepackage{times,latexsym}
\usepackage{url}
\usepackage[table]{xcolor}
\usepackage[T1]{fontenc}

\usepackage[acceptedWithA]{tacl2021v1}

\usepackage{xspace,mfirstuc,tabulary}

\newif\iftaclinstructions
\taclinstructionsfalse 
\iftaclinstructions
\renewcommand{\confidential}{}
\renewcommand{\anonsubtext}{(No author info supplied here, for consistency with
TACL-submission anonymization requirements)}
\newcommand{\instr}
\fi

\iftaclpubformat 

\else

\fi

\usepackage{amsmath}
\usepackage{amssymb}
\usepackage{comment}
\usepackage{stmaryrd}
\usepackage{graphicx}
\usepackage[boldmath]{numprint}
\usepackage{multicol}%
\usepackage{multirow}%
\usepackage{tikz}
\usepackage{tkz-euclide}
\usepackage{subfig}
\usetikzlibrary{shapes,decorations}
\usepackage{relsize}
\usepackage{pgfplotstable}
\usepackage{pgfplots}
\pgfplotsset{compat=newest}
\usepackage{pgfplotstable}
\usepgfplotslibrary{colorbrewer}
\usetikzlibrary{calc}
\usepackage{varwidth}
\usepackage{colortbl}
\usepackage{booktabs}
\usepackage{multicol}
\usepackage{float}
\usepackage{placeins}
\usepackage{cleveref}

\definecolor{color1}{HTML}{1b9e77}
\definecolor{color2}{HTML}{d95f02}
\definecolor{color3}{HTML}{7570b3}
\definecolor{color4}{HTML}{e7298a}

\newcommand{\gradA}{\cellcolor[HTML]{8b0000}\color{white}}
\newcommand{\gradB}{\cellcolor[HTML]{a13301}\color{white}}
\newcommand{\gradC}{\cellcolor[HTML]{b55503}\color{white}}
\newcommand{\gradD}{\cellcolor[HTML]{c7760e}}
\newcommand{\gradE}{\cellcolor[HTML]{d79621}}
\newcommand{\gradF}{\cellcolor[HTML]{e5b737}}
\newcommand{\gradG}{\cellcolor[HTML]{f1d850}}
\newcommand{\gradH}{\cellcolor[HTML]{fafa6e}}
\newcommand{\gradI}{\cellcolor[HTML]{d7e45c}}
\newcommand{\gradJ}{\cellcolor[HTML]{b5ce4b}}
\newcommand{\gradK}{\cellcolor[HTML]{94b93b}}
\newcommand{\gradL}{\cellcolor[HTML]{73a32c}}
\newcommand{\gradM}{\cellcolor[HTML]{538e1d}\color{white}}
\newcommand{\gradN}{\cellcolor[HTML]{32790e}\color{white}}
\newcommand{\gradO}{\cellcolor[HTML]{006400}\color{white}}

\newcommand{\bestCell}[1]{\underline{\textbf{\numprint{#1}}}}

\newcommand{\cmark}{\multicolumn{1}{c}{\color{green!75!black}\ding{51}}}%
\newcommand{\xmark}{\multicolumn{1}{c}{\color{red}\ding{55}}}%

\definecolor{airforceblue}{rgb}{0.36, 0.54, 0.66}

\usepgfplotslibrary{colormaps}
\pgfplotsset{
    colormap={darktrafficlights}{
        rgb255=(139,0,0)
        rgb255=(250,250,110)
        rgb255=(0,100,0)
    },
    colormap name=darktrafficlights,
}

\pgfplotstableset{
    /color cells/min/.initial=0,
    /color cells/max/.initial=1000,
    /color cells/textcolor/.initial=,
    color cells/.code={%
        \pgfqkeys{/color cells}{#1}%
        \pgfkeysalso{%
            postproc cell content/.code={%
                \begingroup
                \pgfkeysgetvalue{/pgfplots/table/@preprocessed cell content}\value
                \ifx\value\empty
                    \endgroup
                \else
                \pgfmathfloatparsenumber{\value}%
                \pgfmathfloattofixed{\pgfmathresult}%
                \let\value=\pgfmathresult
                \pgfplotscolormapaccess
                    [\pgfkeysvalueof{/color cells/min}:\pgfkeysvalueof{/color cells/max}]
                    {\value}
                    {\pgfkeysvalueof{/pgfplots/colormap name}}%
                \pgfkeysgetvalue{/pgfplots/table/@cell content}\typesetvalue%
                \pgfkeysgetvalue{/color cells/textcolor}\textcolorvalue%
                \toks0=\expandafter{\typesetvalue}%
                \xdef\temp{%
                    \noexpand\pgfkeysalso{%
                        @cell content={%
                            \noexpand\cellcolor[rgb]{\pgfmathresult}%
                            \noexpand\definecolor{mapped color}{rgb}{\pgfmathresult}%
                            \ifnum16>\pgfkeysvalueof{/pgfplots/table/@unprocessed cell content}
                                \noexpand\color{white}%
                            \else
                                \ifnum84<\pgfkeysvalueof{/pgfplots/table/@unprocessed cell content}
                                    \noexpand\color{white}%
                                \fi%
                            \fi%
                            \the\toks0%
                        }%
                    }%
                }%
                \endgroup%
                \temp%
                \fi%
            }%
        }%
    }
}

\title{How to Dissect a Muppet: The Structure of Transformer Embedding Spaces}

\author{
  Timothee Mickus\Thanks{The work described in the present paper was conducted chiefly while at ATILF.} \\
  University of Helsinki  \\
  \texttt{timothee.mickus} \\ \texttt{@helsinki.fi}
  \And
  Denis Paperno \\
  Utrecht University \\
  \texttt{d.paperno@uu.nl}
  \And
  Mathieu Constant \\
  Université de Lorraine, CNRS,\\ATILF\\
  \texttt{Mathieu.Constant}\\\texttt{@univ-lorraine.fr}
}

\date{}

\begin{document}
\maketitle
\begin{abstract}
  Pretrained embeddings based on the Transformer architecture have taken the NLP community by storm.
  We show that they can mathematically be reframed as a sum of vector factors and showcase how to use this reframing to study the impact of each component.
  We provide evidence that multi-head attentions and feed-forwards are not equally useful in all downstream applications, as well as a quantitative overview of the effects of finetuning on the overall embedding space.
  This approach allows us to draw connections to a wide range of previous studies, from vector space anisotropy to attention weights.
\end{abstract}
\section{Introduction}
The Transformer architecture \citep{NIPS2017_3f5ee243} has taken the NLP community by storm. 
Based on the attention mechanism \citep{bahdanau-etal-2015-neural,luong-etal-2015-effective}, it was shown to outperform recurrent architectures on a wide variety of tasks.
Another step was taken with pretrained language models derived from this architecture \citep[BERT,][a.o.]{devlin-etal-2019-bert}: they now embody the default approach to a vast swath of NLP applications.
Success breeds scrutiny; likewise the popularity of these models has fostered research in explainable NLP interested in the behavior and explainability of pretrained language models \citep{rogers-etal-2020-primer}.


In this paper, we develop a novel decomposition of Transformer output embeddings.
Our approach consists in quantifying the contribution of each network submodule to the output contextual embedding, and grouping those into four terms: (i) what relates to the input for a given position, (ii) what pertains to feed-forward submodules, (iii) what corresponds to multi-head attention, and (iv) what is due to vector biases.

This allows us to investigate Transformer embeddings without relying on attention weights or treating the entire model as a black box, as is most often done in the literature.
The usefulness of our method is demonstrated on BERT: 
our case study yields enlightening connections to state-of-the-art work on Transformer explainability, evidence that multi-head attentions and feed-forwards are not equally useful in all downstream applications, as well as an overview of the effects of finetuning on the embedding space.
We also provide a simple and intuitive measurement of the importance of any term in this decomposition with respect to the whole embedding.

We will provide insights on the Transformer architecture in \Cref{sec:transformer}, and showcase how these insights can translate into experimental investigations in \Cref{sec:exp-viz,sec:exp-mlm,sec:exp-wsd,sec:exp-ner}.
We will conclude with connections to other relevant works in \Cref{sec:sota} and discuss future perspectives in \Cref{sec:ccl}.

\section{Additive structure in Transformers} \label{sec:transformer}

We show that the Transformer embedding $\mathbf{e}_t$ for a token $t$ is as a sum of four 
\emph{terms}:
\begin{equation} \label{eq:global}
    	\mathbf{e}_t = \mathbf{i}_t + \mathbf{h}_t + \mathbf{f}_t + \mathbf{c}_t
\end{equation}
where $\mathbf{i}_t$ can be thought of as a classical static embedding, $\mathbf{f}_t$ and $\mathbf{h}_t$ are the cumulative contributions at every layer of the feed-forward submodules and the MHAs respectively, and $\mathbf{c}_t$ corresponds to biases accumulated across the model.

\Cref{eq:global} provides interpretable and quantifiable terms that can explain the behavior of specific components of the Transformer architecture.
More precisely, it characterizes what is the impact of adding another sub-layer on top of what was previously computed: the terms in \Cref{eq:global} are defined as sums across (sub-)layers;
hence we can track how a given sublayer transforms its input, and show that this effect can be thought of as adding another vector to a previous sum.
This layer-wise sum of submodule outputs also allows us to provide a first estimate of which parameters are most relevant to the overall embedding space: a submodule whose output is systematically negligible has its parameters set so that its influence on subsequent computations is minimal.

The formulation in \Cref{eq:global} more generally relies on the additive structure of Transformer embedding spaces. 
We start by reviewing the Transformer architecture in \Cref{{subsec:architecture}}, before discussing our decomposition in greater detail in \Cref{subsec:decomp} and known limitations in \Cref{subsec:limits}.

\subsection{Transformer encoder architecture} \label{subsec:architecture}

\begin{table}
    \centering
    \begin{tabular}{| p{0.1\linewidth} p{0.8\linewidth} |}
        \hline
        $\mathbf{A}$ & matrix \\
        $\left(\mathbf{A}\right)_{t,\cdot}$ & $t$\textsuperscript{th} row of $\mathbf{A}$ \\
        $\mathbf{a}$ & (row) vector \\ 
        $a$, $\alpha$ & scalars \\
        $\mathbf{W}^{(\text{M})}$ & item linked to submodule $M$ \\ 
        $\mathbf{a}\oplus\mathbf{b}$ & Concatenation of vectors $\mathbf{a}$ and $\mathbf{b}$ \\ 
        $\bigoplus\limits_n \mathbf{a}_n$ & $\mathbf{a}_1\oplus\mathbf{a}_2\oplus\dots\oplus\mathbf{a}_n$\\
        $\mathbf{a}\odot\mathbf{b}$ & Element-wise multiplication of $\mathbf{a}$ and $\mathbf{b}$\\
        $\bigodot\limits_n \mathbf{a}_n$ & $\mathbf{a}_1\odot\mathbf{a}_2\odot\dots\odot\mathbf{a}_n$\\
        $\vec{1}$ & vector with all components set to 1 \\
        $\mathbf{0}_{m,n}$ & null matrix of shape $m \times n$ \\
        $\mathbf{I}_n$ & Identity matrix of shape $n \times n$ \\ \hline
    \end{tabular}
    \caption{Notation}
    \label{tab:notations}
\end{table}

\begin{figure}
    \centering
    \resizebox{\columnwidth}{!}{\begin{tikzpicture}[
              font=\sf \scriptsize,
              >=LaTeX,
              cell/.style={
                  rectangle,
                  rounded corners=5mm,
                  very thick,
                  },
              operator/.style={
                  circle,
                  draw=black,
                  fill=white,
                  thick,
                  inner sep=-0.5pt,
                  minimum height =.2cm,
                  },
              function/.style={
                  ellipse,
                  draw=black, 
                  fill=white,
                  thick,
                  inner sep=1pt
                  },
              ct/.style={
                  circle,
                  draw,
                  line width = .75pt,
                  minimum width=1cm,
                  inner sep=1pt,
                  },
              gt/.style={
                  rectangle,
                  draw,
                  minimum width=4mm,
                  minimum height=3mm,
                  inner sep=1pt
                  },
              mylabel/.style={
                  font=\scriptsize\sffamily
                  },
              ArrowC1/.style={
                  rounded corners=.25cm,
                  thick,
                  },
              ArrowC2/.style={
                  rounded corners=.5cm,
                  thick,
                  },
              ]

              \foreach \layer [count=\lidx from 0] in {1, l, L}{
                \node [cell, minimum width=2.5cm, minimum height=6.5cm, fill=orange!50!white, draw=orange!95!black] at (3 * \lidx, 0.65){} ;
                \node at (0.33 + 3 * \lidx, -2.5) {\smaller \it layer $\layer$};
                \foreach \slfunc [count=\slidx from 0] in {{MHA},{FF}} {
                    \node [cell, minimum width=2cm, minimum height=2.75cm, fill=airforceblue!50!white, draw=airforceblue!95!black] at (3 * \lidx, -0.85 + \slidx * 3.125){} ;
                    \node [function] (f\slfunc\lidx) at (3 * \lidx + 0.5, -1.5 + \slidx * 3.25) {\smaller \slfunc};%
                    \node [operator] (skip\slfunc\lidx) at (3 * \lidx - 0.33, -0.75 + \slidx * 3.25) {+};%
                    \node [function] (ln\slfunc\lidx) at (3 * \lidx - 0.33, 0.125 + \slidx * 3.25) {\smaller LN};%
                    \draw [->, ArrowC1] (f\slfunc\lidx.north) |- (skip\slfunc\lidx.east);
                    \draw [->, ArrowC1] (skip\slfunc\lidx.north) -- (ln\slfunc\lidx.south);
                    
                }
                \node [gt] (out\lidx) at (3 * \lidx + 0.5, 4.25) {$\mathbf{y}_{\layer,t}^{}$};
                \draw [->, ArrowC1] (lnMHA\lidx.north) -- (skipFF\lidx.south);
                \draw [->, ArrowC1] (lnMHA\lidx.north) |- (fFF\lidx.west);
                \draw [->, ArrowC1] (lnFF\lidx.north) |- (out\lidx.west);
              }
              \foreach \nextlidx [count=\prevlidx from 0] in {1, 2}{

                \node (interrupt\nextlidx) at (3 * \prevlidx + 1.5, 0.875) {$\vdots$};
                \node (toskip\nextlidx) at (3 * \prevlidx + 2,  -1.5) {};

                \draw [ArrowC1] (out\prevlidx.east) -| (interrupt\nextlidx.north);
                \draw [->, ArrowC1]  (interrupt\nextlidx.south) |- (fMHA\nextlidx.west) ;
                \draw [->, ArrowC1]  (toskip\nextlidx.east) -| (skipMHA\nextlidx.south) ;
                
              }
              
              \node [gt] (input) at (-.33, -3.125) {$\mathbf{x}_{0,t}$};
              \draw [->, ArrowC1] (input.north) -- (skipMHA0.south);
              \draw [->, ArrowC1] (input.north) |- (fMHA0.west);

            \node at (0.25 , -2.) {\smaller \it sublayer $1$};
            \node at (0.25 +3 , -2.) {\smaller \it subl. $2l-1$};
            \node at (0.25 +6 , -2.) {\smaller \it subl. $2L-1$};
            \node at (0.25 , -2.+3.125) {\smaller \it sublayer 2};
            \node at (0.25 +3 , -2.+3.125) {\smaller \it subl. $2l$};
            \node at (0.25 +6 , -2.+3.125) {\smaller \it subl. $2L$};

          \end{tikzpicture}}
    \caption{Overview of a Transformer encoder.}
    \label{fig:tf-arch}
\end{figure}
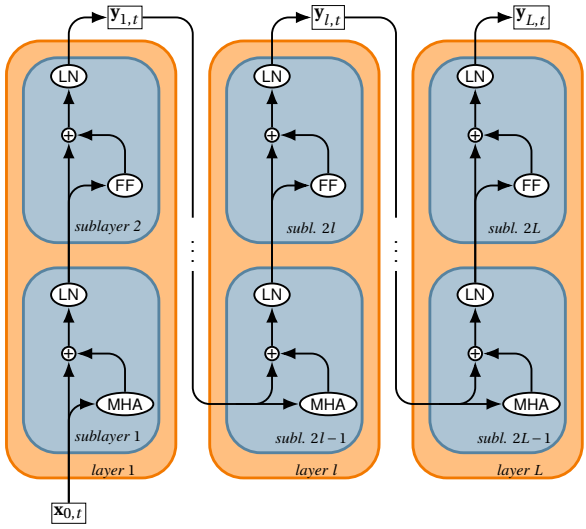

Let's start by characterizing the Transformer architecture of \citet{NIPS2017_3f5ee243} in the notation described in \Cref{tab:notations}.

Transformers are often defined using three hyperparameters: the number of layers $L$, the dimensionality of the hidden representations $d$, and $H$, the number of attention heads in multi-head attentions.
Formally, a Transformer model is a stack of \emph{sublayers}.
A visual representation is shown in \Cref{fig:tf-arch}.
Two sublayers are stacked to form a single Transformer \emph{layer}: the first corresponds to a MHA, and the second to a FF.
A Transformer with $L$ layers contains $\Lambda=2L$ sublayers.
In \Cref{fig:tf-arch} two sublayers (in blue) are grouped into one layer, and $L$ layers are stacked one after the other.

Each sublayer is centered around a specific sublayer function.
Sublayer functions map an input $\mathbf{x}$ to an output $\mathbf{y}$, and can either be \emph{feed-forward} submodules or \emph{multi-head attention} submodules.

Feed-forwards (FFs) are subnets of the form:
\begin{equation*}
    \mathbf{y}_{t}^{(\text{FF})} = \phi \left( \mathbf{x}_t \mathbf{W}^{(\text{FF,I})}_{}  + \mathbf{b}^{(\text{FF,I})}_{} \right) \mathbf{W}^{(\text{FF,O})}_{}  + \mathbf{b}^{(\text{FF,O})}_{}
\end{equation*}
where $\phi$ is a non-linear function, such as {ReLU} or {GELU} \citep{hendrycks-2016-gelu}.
Here, ${}^{(\dots,\text{I})}$ and ${}^{(\dots,\text{O})}$ distinguish the input and output linear projections, whereas the index $t$ corresponds to the token position.
Input and output dimensions are equal, whereas the intermediary layer dimension (i.e., the size of the hidden representations to which the non-linear function $\phi$ will be applied) is larger, typically of $b=1024$ or 2048. 
In other words,  $ \mathbf{W}^{(\text{FF,I})}$ is of shape $d \times b$, $\mathbf{b}^{(\text{FF,I})}$ of size $b$, $\mathbf{W}^{(\text{FF,O})}$ is of shape $k \times d$, and $\mathbf{b}^{(\text{FF,O})}$ of size $d$.

Multi-head attention mechanisms (MHAs) are concatenations of scaled-dot \emph{attention heads}:

\begin{equation*}
    \mathbf{y}^{(\text{MHA})}_{t} = \left(\bigoplus\limits_{h=1}^H (\mathbf{A}_{h})_{t,\cdot}\right)  \mathbf{W}^{(\text{MHA,O})} + \mathbf{b}^{(\text{MHA,O})}\\
\end{equation*}
{\color{black} where $(\mathbf{A}_{h})_{t,\cdot}$ is the $t$\textsuperscript{th} row vector  of the following $n \times d/H$ matrix $\mathbf{A}_h$:}
\begin{equation*}
    \mathbf{A}_{h} = \text{softmax} \left( \frac{ \mathbf{Q}_{h}^{}\mathbf{K}_h^T}{\sqrt{d/H}} \right)  \mathbf{V}_h
\end{equation*}
with $h$ an index tracking attention heads.
The parameters $\mathbf{W}^{(\text{MHA,O})}$ of shape $d\times d$, $\mathbf{b}^{(\text{MHA,O})}$ of size $d$, and the queries $\mathbf{Q}_h$,  keys $\mathbf{K}_h$ and values $\mathbf{V}_h$ are simple linear projections of shape $n \times (d/H)$, computed from all inputs $\mathbf{x}_1, ~ \dots, ~ \mathbf{x}_n$: 
\begin{align*}
    (\mathbf{Q}_h)_{t,\cdot} &= \mathbf{x}_{t}\mathbf{W}_{h}^{(\text{Q})} +\mathbf{b}_{h}^{(\text{Q})} \\
    (\mathbf{K}_h)_{t,\cdot} &= \mathbf{x}_{t}\mathbf{W}_{h}^{(\text{K})} +\mathbf{b}_{h}^{(\text{K})} \\
    (\mathbf{V}_h)_{t,\cdot} &= \mathbf{x}_{t}\mathbf{W}_{h}^{(\text{V})} +\mathbf{b}_{h}^{(\text{V})} 
    \end{align*}
where the weight matrices $\mathbf{W}_{h}^{(\text{Q})}$, $\mathbf{W}_{h}^{(\text{K})}$ and $\mathbf{W}_{h}^{(\text{V})}$ are of the shape $d \times (d/H) $, with $H$ the number of attention heads, and biases $\mathbf{b}_{h}^{(\text{Q})}$, $\mathbf{b}_{h}^{(\text{K})}$  and $\mathbf{b}_{h}^{(\text{V})}$ are of size $d/H$. 
This component is often analyzed in terms of \emph{attention weights} $\alpha_{h}$, which correspond to the softmax dot-product between keys and queries.
In other words, the product $\text{softmax} (\mathbf{Q}_h^{} \mathbf{K}_h^T / \sqrt{d/H})$ can be thought of as $n \times n$ matrix of weights in an average over the transformed input vectors $\mathbf{x}_{t'} \mathbf{W}_{h}^{(\text{V})}  + \mathbf{b}_{h}^{(\text{V})}$ \citep[eqs. (1) to (4)]{kobayashi-etal-2020-attention}: multiplying these weights with the value projection $\mathbf{V}_h$ yields a weighted sum of value projections:
\begin{equation*}
    (\mathbf{A}_{h})_{t,\cdot} = \sum\limits_{t'=1}^n \alpha_{h,t,t'} (\mathbf{V}_{h})_{t',\cdot}
\end{equation*}
where $\alpha_{h,t,t'}$ is the component at row $t$ and column $t'$ of this attention weights matrix.

Lastly, after each sublayer function $S$, a \emph{residual connection} and a \emph{layer normalization} \citep[LN,][]{ba2016layer} are applied:
\begin{equation*}
    \mathbf{y}_{t}^{(\text{LN})} = \mathbf{g} \odot \left[\frac{S\left(\mathbf{x}_t\right) + \mathbf{x}_t - \left(m_t \cdot \vec{1}\right)}{s_t}\right] + \mathbf{b}^{(\text{LN})}
\end{equation*}
The gain $\mathbf{g}$ and bias $\mathbf{b}^{(\text{LN})}$ are learned parameters with $d$ components each;
$m_t \cdot \vec{1} $  is the vector $(1,~\cdots,~1)$  scaled by the mean component value $m_t$ of the input vector $S\left(\mathbf{x}_t\right) + \mathbf{x}_t$; $s_t$ is the standard deviation of the component values of this input.
As such, a LN performs a $z$-scaling, followed by the application of the gain $\mathbf{g}$ and the bias $\mathbf{b}^{(\text{LN})}$.

%
To kick-start computations, a sequence of static vector representations $\mathbf{x}_{0,1} \dots \mathbf{x}_{0,n}$ with $d$ components each is fed into the first layer. 
This initial input corresponds to the sum of a static lookup word embedding and a positional encoding.\footnote{
    In BERT \citet{devlin-etal-2019-bert}, additional terms to this static input encode the segment a token belongs to, and a LN is added before the very first sublayer.
    Other variants also encode positions by means of an offset in attention heads \citep{huang-etal-2018-improved,shaw-etal-2018-self}. 
}
\subsection{Mathematical re-framing} \label{subsec:decomp}
We now turn to the decomposition proposed in \Cref{eq:global}: $\mathbf{e}_t = \mathbf{i}_t + \mathbf{f}_t + \mathbf{h}_t + \mathbf{c}_t$.\footnote{
    We empirically verified that components from attested embeddings $\mathbf{e}_t$ and those derived from \cref{eq:global} are systematically equal up to $\pm 10^{-7}$.
} 
We provide a derivation in Appendix~\ref{sec:appendix-deriv}.


The term $\mathbf{i}_t$ corresponds to the input embedding (i.e., the positional encoding, the input word-type embedding, and the segment encoding in BERT-like models), after having gone through all the LN gains and rescaling:
\begin{equation} \label{eq:i-term}
    \mathbf{i}_t = \frac{\bigodot\limits_{\lambda=1}^{\Lambda} \mathbf{g}_\lambda^{} }{\prod\limits_{\lambda=1}^{\Lambda} s_{\lambda,t}} \odot \mathbf{x}_{0,t}
\end{equation}
where $\Lambda=2L$ ranges over all sublayers.
Here, the $\mathbf{g}_\lambda$ correspond to the learned gain parameters of the LNs, whereas the $s_{\lambda,t}$ scalar derive from the $z$-scaling performed in the $\lambda$\textsuperscript{th} LN, as defined above.
The input $\mathbf{x}_{0,t}$ consists of the sum of a static lookup embedding and a positional encoding: as such, it resembles an uncontextualized embedding. 

The next two terms capture the outputs of specific submodules, either FFs or MHAs. 
As such, their importance and usefulness will differ from task to task.
The term $\mathbf{f}_t$ is the sum of the outputs of the FF submodules. 
Submodule outputs pass through LNs of all the layers above, hence: 
\begin{equation}  \label{eq:f-term}
    \mathbf{f}_t = \sum_{l=1}^L \frac{\bigodot\limits_{\lambda=2l}^{\Lambda} \mathbf{g}_\lambda^{} }{\prod\limits_{\lambda=2l}^{\Lambda} s_{\lambda,t}} \odot  \Tilde{\mathbf{f}}_{l,t}
\end{equation}
where $\Tilde{\mathbf{f}}_{l,t} = \phi \left( \mathbf{x}_{l,t}^{(\text{FF})}  \mathbf{W}^{(\text{FF,I})}_l+ \mathbf{b}^{(\text{FF,I})}_l \right) \mathbf{W}^{(\text{FF,O})}_l $ is the unbiased output at the position $t$ of the FF submodule for this layer $l$.

The term $\mathbf{h}_t$ corresponds to the sum across layers of each MHA output, having passed through the relevant LNs.
As MHAs are entirely linear, we can further describe each output as a sum over all $H$ heads of a weighted bag-of-words of the input representations to that submodule.
Or: 
\begin{align}  \label{eq:h-term}
    \mathbf{h}_t &=  \sum_{l=1}^L \left( \frac{\bigodot\limits_{\lambda=2l-1}^{\Lambda} \mathbf{g}_\lambda^{}}{\prod\limits_{\lambda=2l-1}^{\Lambda} s_{\lambda,t}}  \odot \left[\sum_{h=1}^H \sum_{t'=1}^n \alpha_{l,h,t,t'} \mathbf{x}_{l,t'} \mathbf{Z}_{l,h} \right]\right) \nonumber \\
    \mathbf{Z}_{l,h} &=  \mathbf{W}_{l,h}^{(\text{V})} \mathbf{M}_h^{} \mathbf{W}^{(\text{MHA,O})}_l
\end{align}
where $\mathbf{Z}_{l,h}$ 
corresponds to passing an input embedding through the unbiased values projection $\mathbf{W}_{l,h}^{(\text{V})}$ of the head $h$, then projecting it from a $d/H$-dimensional subspace onto a $d$-dimensional space using a zero-padded identity matrix:
\begin{equation*}
    \mathbf{M}_h^{} = \begin{bmatrix}
        \mathbf{0}_{d/H,(h-1) \times d/H} & \mathbf{I}_{d/H} & \mathbf{0}_{d/H,(H-h) \times d/H} 
    \end{bmatrix}    
\end{equation*}
and finally passing it through the  unbiased outer projection $\mathbf{W}^{(\text{MHA,O})}_l$ of the relevant MHA.

In the last term $\mathbf{c}_t$, we collect all the biases. 
We don't expect these offsets to be meaningful but rather to depict a side-effect of the architecture:
\begin{align}  \label{eq:c-term}
	\mathbf{c}_t =& \sum_{\lambda=1}^\Lambda \left( \frac{\bigodot\limits_{\lambda'=\lambda+1}^{\Lambda} \mathbf{g}_{\lambda'}}{\prod\limits_{\lambda'=\lambda+1}^{\Lambda} s_{\lambda',t}}  \odot  \mathbf{b}_\lambda^{(\text{\smaller LN})} %
	              - \frac{\bigodot\limits_{\lambda'=\lambda}^{\Lambda} \mathbf{g}_{\lambda'}}{\prod\limits_{\lambda'=\lambda}^{\Lambda} s_t^{\lambda'}} \odot \left(m_{\lambda,t} \cdot \vec{1}\right) \right) \nonumber \\ %
	                &+ \sum_{l=1}^L\frac{\bigodot\limits_{\lambda=2l-1}^{\Lambda} \mathbf{g}_\lambda}{\prod\limits_{\lambda=2l-1}^{\Lambda} s_{\lambda,t}} \odot \left[\mathbf{b}_l^{(\text{\smaller MHA,O})}  +\left(\bigoplus\limits_{h=1}^H \mathbf{b}_{l,h}^{(\text{\smaller  V})}\right) \mathbf{W}^{(\text{\smaller MHA,O})}_l  \right]  \nonumber \\ %
                      &+ \sum_{l=1}^L\frac{\bigodot\limits_{\lambda=2l}^{\Lambda} \mathbf{g}_\lambda}{\prod\limits_{\lambda=2l}^{\Lambda} s_{\lambda,t}} \odot \mathbf{b}_l^{(\text{\smaller FF,O})} %
\end{align}
The concatenation $\bigoplus_h^{} \mathbf{b}_{l,h}^{(\text{V})}$ here is equivalent to a sum of zero-padded identity matrices: $\sum_h^{} \mathbf{b}_{l,h}^{(\text{V})} \mathbf{M}_h$.
This term $\mathbf{c}_t$ includes the biases $\mathbf{b}_\lambda^{(\text{LN})}$ and mean-shifts $m_{\lambda,t} \cdot \vec{1}$ of the LNs, the outer projection biases of the FF submodules $\mathbf{b}_l^{(\text{FF,O})}$, the outer projection bias in each MHA submodule $\mathbf{b}_l^{(\text{MHA,O})}$ and the value projection biases, mapped through the outer MHA projection $\left(\bigoplus_{h} \mathbf{b}_{l,h}^{(\text{V})}\right)\mathbf{W}^{(\text{MHA,O})}_l $.
\footnote{
    In the case of relative positional embeddings applied to value projections \citep{shaw-etal-2018-self}, it is rather straightforward to follow the same logic so as to include relative positional offset in the most appropriate term.
}

\subsection{Limitations of \Cref{eq:global}}
\label{subsec:limits}

The decomposition proposed in \Cref{eq:global} comes with a few caveats that are worth addressing explicitly.
Most importantly, \Cref{eq:global} does not entail that the terms are independent from one another.
For instance, the scaling factor $1/\prod s_{\lambda,t}$ systematically depends on the magnitude of earlier hidden representations.
\Cref{eq:global} only stresses that a Transformer embedding can be decomposed as a sum of the outputs of its submodules: it does not fully disentangle computations. 
%
We leave the precise definition of computation disentanglement and its elaboration for the Transformer to future research, and focus here on the decomposition proposed in \Cref{eq:global}

In all, the major issue at hand is the $\mathbf{f}_t$ term: it is the only term that cannot be derived as a linear composition of vectors, due to the non-linear function used in the FFs.
Aside from the $\mathbf{f}_t$ term, non-linear computations all devolve into scalar corrections (namely the LN $z$-scaling factors $s_{\lambda,t}$ and $m_{\lambda,t}$ and the attention weights $\alpha_{l,h}$).
As such, $\mathbf{f}_t$ is the single bottleneck that prevents us from entirely decomposing a Transformer embedding as a linear combination of sub-terms.
     
As the non-linear functions used in Transformers are generally either ReLU or GELU, which both behave almost linearly for a high enough input value, it is in principle possible that the FF submodules can be approximated by a purely linear transformation, depending on the exact set of parameters they converged onto.
It is worth assessing this possibility.
Here, we learn a least-square linear regression mapping the $z$-scaled inputs of every FF to its corresponding $z$-scaled output. 
We use the BERT base uncased model of \citet{devlin-etal-2019-bert} and a random sample of 10,000 sentences from the Europarl English section \citep{koehn-2005-europarl}, or almost 900,000 word-piece tokens, and fit the regressions using all 900,000 embeddings. 

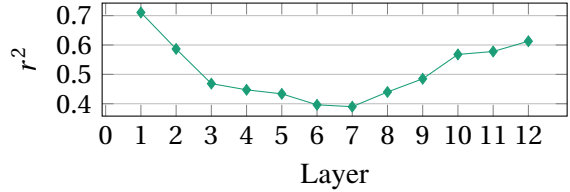
\begin{figure}[thbp]
    \centering
    \begin{tikzpicture}
\pgfplotstableread[col sep=comma,]{data/xlayers.csv}\xlayerdata;

\begin{axis}[
  xlabel={Layer},
  xtick={0,1,...,12},
  ymajorgrids,
  width=\linewidth,
  height=0.4\linewidth,
  ylabel={$r^2$},
  legend style={at={(1,1)},anchor=north east, legend columns=4,},
  cycle list/Dark2,
]

\addplot+[
mark=diamond*,
]
table [x=layer, y=ffr2, col sep=comma] {\xlayerdata};


\end{axis}
\end{tikzpicture}
    \caption{Fitting the $\mathbf{f}_t$ term:  $r^2$ across layers}
    \label{fig:ffr2}
\end{figure}

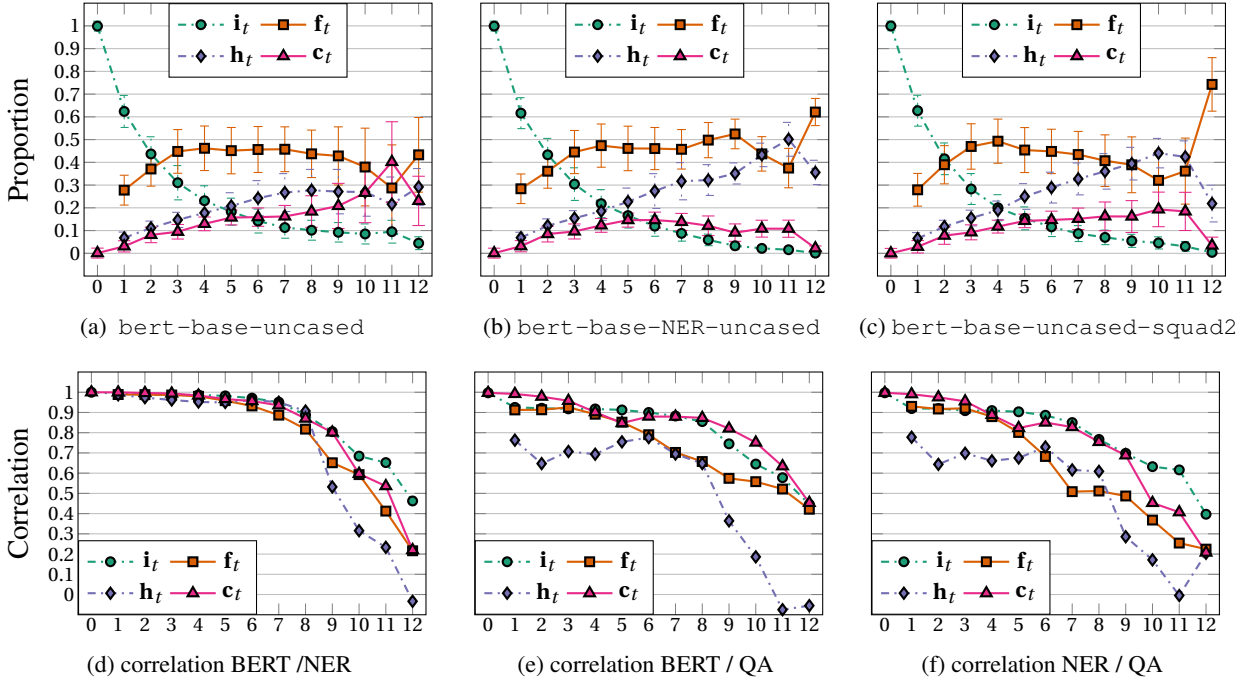
\begin{figure*}
    \centering
    \subfloat[\label{fig:breakdown:base} \tt bert-base-uncased]{
        \begin{tikzpicture}
\pgfplotscreateplotcyclelist{mycolorlisttest}{%
    color1,thick,dash dot,every mark/.append style={solid,draw=black,fill=color1},mark=*,mark size=1.66pt\\
    color2,thick,every mark/.append style={draw=black,fill=color2},mark=square*,mark size=1.66pt\\
    color3,thick,dash dot,every mark/.append style={solid,draw=black,fill=color3},mark=diamond*,mark size=2.1pt\\
    color4,thick,every mark/.append style={draw=black,fill=color4},mark=triangle*,mark size=2.33pt\\
}
\pgfplotstableread[col sep=comma,]{data/xlayers.csv}\xlayerdata;
\pgfplotsset{every x tick label/.append style={font=\scriptsize, yshift=0.5ex}}
\pgfplotsset{every y tick label/.append style={font=\scriptsize, xshift=0.5ex}}
\begin{axis}[
  xmax=12.5, xmin=-0.5, ymax=1.1, ymin=-0.1,  
  xtick={0,1,...,12},
  ytick={0.0,0.1,...,1.0},
  ymajorgrids,
  width=0.385\linewidth,
  height=0.325\linewidth,
  ylabel={Proportion},
  legend style={at={(0.5,1)},anchor=north, legend columns=2,},
  cycle list name=mycolorlisttest,
]
\addplot+[
error bars/.cd,
    y explicit,
    y dir=both,
]
table [x=layer, y=ipt, y error=ipterr,] {\xlayerdata};
\addlegendentry{\smaller$\mathbf{i}_t$};
\addplot+[
error bars/.cd,
    y explicit,
    y dir=both,
]
table [x=layer, y=ff, col sep=comma, y error=fferr] {\xlayerdata};
\addlegendentry{\smaller$\mathbf{f}_t$};
\addplot+[
error bars/.cd,
    y explicit,
    y dir=both,
]
table [x=layer, y=mha, col sep=comma,y error=mhaerr] {\xlayerdata};
\addlegendentry{\smaller$\mathbf{h}_t$};
\addplot+[
error bars/.cd,
    y explicit,
    y dir=both,
]
table [x=layer, y=norm, col sep=comma,y error=normerr] {\xlayerdata};
\addlegendentry{\smaller$\mathbf{c}_t$};
\end{axis}
\end{tikzpicture}
    }
    \subfloat[\label{fig:breakdown:ner}\tt bert-base-NER-uncased]{
        \begin{tikzpicture}
\pgfplotscreateplotcyclelist{mycolorlisttest}{%
    color1,thick,dash dot,every mark/.append style={solid,draw=black,fill=color1},mark=*,mark size=1.66pt\\
    color2,thick,every mark/.append style={draw=black,fill=color2},mark=square*,mark size=1.66pt\\
    color3,thick,dash dot,every mark/.append style={solid,draw=black,fill=color3},mark=diamond*,mark size=2.1pt\\
    color4,thick,every mark/.append style={draw=black,fill=color4},mark=triangle*,mark size=2.33pt\\
}
\pgfplotstableread[col sep=comma,]{data/xlayers-tuned.csv}\xlayerdata;
\pgfplotsset{every x tick label/.append style={font=\scriptsize, yshift=0.5ex}}
\begin{axis}[
  xmax=12.5, xmin=-0.5, ymax=1.1, ymin=-0.1,  
  xtick={0,1,...,12},
  ytick={0.0,0.1,...,1.0},
  yticklabels={},
  ymajorgrids,
  width=0.385\linewidth,
  height=0.325\linewidth,
  legend style={at={(0.5,1)},anchor=north, legend columns=2,},
  cycle list name=mycolorlisttest,
]
\addplot+[
error bars/.cd,
    y explicit,
    y dir=both,
]
table [x=layer, y=ipt, y error=ipterr,] {\xlayerdata};
\addlegendentry{\smaller$\mathbf{i}_t$};
\addplot+[
error bars/.cd,
    y explicit,
    y dir=both,
]
table [x=layer, y=ff, col sep=comma, y error=fferr] {\xlayerdata};
\addlegendentry{\smaller$\mathbf{f}_t$};
\addplot+[
error bars/.cd,
    y explicit,
    y dir=both,
]
table [x=layer, y=mha, col sep=comma,y error=mhaerr] {\xlayerdata};
\addlegendentry{\smaller$\mathbf{h}_t$};
\addplot+[
error bars/.cd,
    y explicit,
    y dir=both,
]
table [x=layer, y=norm, col sep=comma,y error=normerr] {\xlayerdata};
\addlegendentry{\smaller$\mathbf{c}_t$};
\end{axis}
\end{tikzpicture}
    }
    \subfloat[\label{fig:breakdown:squad}\tt bert-base-uncased-squad2]{
        \begin{tikzpicture}
\pgfplotscreateplotcyclelist{mycolorlisttest}{%
    color1,thick,dash dot,every mark/.append style={solid,draw=black,fill=color1},mark=*,mark size=1.66pt\\
    color2,thick,every mark/.append style={draw=black,fill=color2},mark=square*,mark size=1.66pt\\
    color3,thick,dash dot,every mark/.append style={solid,draw=black,fill=color3},mark=diamond*,mark size=2.1pt\\
    color4,thick,every mark/.append style={draw=black,fill=color4},mark=triangle*,mark size=2.33pt\\
}
\pgfplotstableread[col sep=comma,]{data/xlayers-squad.csv}\xlayerdata;
\pgfplotsset{every x tick label/.append style={font=\scriptsize, yshift=0.5ex}}
\begin{axis}[
  xmax=12.5, xmin=-0.5, ymax=1.1, ymin=-0.1,  
  xtick={0,1,...,12},
  ytick={0.0,0.1,...,1.0},
  yticklabels={},
  ymajorgrids,
  width=0.385\linewidth,
  height=0.325\linewidth,
  legend style={at={(0.5,1)},anchor=north, legend columns=2,},
  cycle list name=mycolorlisttest,
]
\addplot+[
error bars/.cd, 
    y explicit,
    y dir=both,
]
table [x=layer, y=ipt, y error=ipterr,] {\xlayerdata};
\addlegendentry{\smaller$\mathbf{i}_t$};
\addplot+[
error bars/.cd,
    y explicit,
    y dir=both,
]
table [x=layer, y=ff, col sep=comma, y error=fferr] {\xlayerdata};
\addlegendentry{\smaller$\mathbf{f}_t$};
\addplot+[
error bars/.cd,
    y explicit,
    y dir=both,
]
table [x=layer, y=mha, col sep=comma,y error=mhaerr] {\xlayerdata};
\addlegendentry{\smaller$\mathbf{h}_t$};
\addplot+[
error bars/.cd,
    y explicit,
    y dir=both,
]
table [x=layer, y=norm, col sep=comma,y error=normerr] {\xlayerdata};
\addlegendentry{\smaller$\mathbf{c}_t$};
\end{axis}
\end{tikzpicture}
    }
    
    \subfloat[\label{fig:correl:base-ner}correlation BERT /NER]{
        \begin{tikzpicture}
\pgfplotscreateplotcyclelist{mycolorlisttest}{%
    color1,thick,dash dot,every mark/.append style={solid,draw=black,fill=color1},mark=*,mark size=1.66pt\\
    color2,thick,every mark/.append style={draw=black,fill=color2},mark=square*,mark size=1.66pt\\
    color3,thick,dash dot,every mark/.append style={solid,draw=black,fill=color3},mark=diamond*,mark size=2.1pt\\
    color4,thick,every mark/.append style={draw=black,fill=color4},mark=triangle*,mark size=2.33pt\\
}
\pgfplotstableread[col sep=space,]{data/correl.csv}\correldata;
\pgfplotsset{every x tick label/.append style={font=\scriptsize, yshift=0.5ex}}
\pgfplotsset{every y tick label/.append style={font=\scriptsize, xshift=0.5ex}}
\begin{axis}[
  xmax=12.5, xmin=-0.5, ymax=1.1, ymin=-0.1,  
  xtick={0,1,...,12},
  ytick={0.0,0.1,...,1.0},
  ymajorgrids,
  width=0.385\linewidth,
  height=0.3\linewidth,
  ylabel={Correlation},
  legend style={at={(0,0)},anchor=south west, legend columns=2,},
  cycle list name=mycolorlisttest,
]
\addplot+[
]
table [x=layer, y=isr] {\correldata};
\addlegendentry{\smaller$\mathbf{i}_t$};
\addplot+[
]
table [x=layer, y=fsr, col sep=space] {\correldata};
\addlegendentry{\smaller$\mathbf{f}_t$};
\addplot+[
]
table [x=layer, y=hsr, col sep=space] {\correldata};
\addlegendentry{\smaller$\mathbf{h}_t$};
\addplot+[
]
table [x=layer, y=csr, col sep=space] {\correldata};
\addlegendentry{\smaller$\mathbf{c}_t$};
\end{axis}
\end{tikzpicture}
    }
    \subfloat[\label{fig:correl:base-squad}correlation BERT / QA]{
        \begin{tikzpicture}
\pgfplotscreateplotcyclelist{mycolorlisttest}{%
    color1,thick,dash dot,every mark/.append style={solid,draw=black,fill=color1},mark=*,mark size=1.66pt\\
    color2,thick,every mark/.append style={draw=black,fill=color2},mark=square*,mark size=1.66pt\\
    color3,thick,dash dot,every mark/.append style={solid,draw=black,fill=color3},mark=diamond*,mark size=2.1pt\\
    color4,thick,every mark/.append style={draw=black,fill=color4},mark=triangle*,mark size=2.33pt\\
}
\pgfplotstableread[col sep=space,]{data/correl-squad.csv}\correldata;
\pgfplotsset{every x tick label/.append style={font=\scriptsize, yshift=0.5ex}}
\begin{axis}[
  xmax=12.5, xmin=-0.5, ymax=1.1, ymin=-0.1,  
  xtick={0,1,...,12},
  ytick={0.0,0.1,...,1.0},
  yticklabels={},
  ymajorgrids,
  width=0.385\linewidth,
  height=0.3\linewidth,
  legend style={at={(0,0)},anchor=south west, legend columns=2,},
  cycle list name=mycolorlisttest,
]
\addplot+[
]
table [x=layer, y=isr] {\correldata};
\addlegendentry{\smaller$\mathbf{i}_t$};
\addplot+[
]
table [x=layer, y=fsr, col sep=space] {\correldata};
\addlegendentry{\smaller$\mathbf{f}_t$};
\addplot+[
]
table [x=layer, y=hsr, col sep=space] {\correldata};
\addlegendentry{\smaller$\mathbf{h}_t$};
\addplot+[
]
table [x=layer, y=csr, col sep=space] {\correldata};
\addlegendentry{\smaller$\mathbf{c}_t$};
\end{axis}
\end{tikzpicture}
    }
    \subfloat[\label{fig:correl:ner-squad}correlation NER / QA]{
        \begin{tikzpicture}
\pgfplotscreateplotcyclelist{mycolorlisttest}{%
    color1,thick,dash dot,every mark/.append style={solid,draw=black,fill=color1},mark=*,mark size=1.66pt\\
    color2,thick,every mark/.append style={draw=black,fill=color2},mark=square*,mark size=1.66pt\\
    color3,thick,dash dot,every mark/.append style={solid,draw=black,fill=color3},mark=diamond*,mark size=2.1pt\\
    color4,thick,every mark/.append style={draw=black,fill=color4},mark=triangle*,mark size=2.33pt\\
}
\pgfplotstableread[col sep=space,]{data/correl-ner-squad.csv}\correldata;
\pgfplotsset{every x tick label/.append style={font=\scriptsize, yshift=0.5ex}}
\begin{axis}[
  xmax=12.5, xmin=-0.5, ymax=1.1, ymin=-0.1,  
  xtick={0,1,...,12},
  ytick={0.0,0.1,...,1.0},
  yticklabels={},
  ymajorgrids,
  width=0.385\linewidth,
  height=0.3\linewidth,
  legend style={at={(0,0)},anchor=south west, legend columns=2,},
  cycle list name=mycolorlisttest,
]
\addplot+[
]
table [x=layer, y=isr] {\correldata};
\addlegendentry{\smaller$\mathbf{i}_t$};
\addplot+[
]
table [x=layer, y=fsr, col sep=space] {\correldata};
\addlegendentry{\smaller$\mathbf{f}_t$};
\addplot+[
]
table [x=layer, y=hsr, col sep=space] {\correldata};
\addlegendentry{\smaller$\mathbf{h}_t$};
\addplot+[
]
table [x=layer, y=csr, col sep=space] {\correldata};
\addlegendentry{\smaller$\mathbf{c}_t$};
\end{axis}
\end{tikzpicture}
    }
    \caption{Relative importance of main terms}
    \label{fig:breakdown}
\end{figure*}

\Cref{fig:ffr2} displays the quality of these linear approximations, as measured by a $r^2$ score. 
We see some variation across layers but never observe a perfect fit: 
30 to 60\% of the observed variance is not explained by a linear map, suggesting BERT actively exploits the non-linearity. 
That the model doesn't simply circumvent the non-linear function to adopt a linear behavior intuitively makes sense:
adding the feed-forward terms is what prevents the model from devolving into a sum of bag-of-words and static embeddings.
While such approaches have been successful \cite{mikolov-etal-2013-linguistic,mitchell-lapata-2010-composition}, a non-linearity ought to make the model more expressive.

In all, the sanity check in \Cref{fig:ffr2} highlights that the interpretation of the $\mathbf{f}_t$ term is the major ``black box'' unanalyzable component remaining under \Cref{eq:global}.
As such, the recent interest in analyzing these modules \citep[e.g.]{geva-etal-2021-transformer,9533563,geva-etal-2022-transformer} is likely to have direct implications for the relevance of the present work.
When adopting the linear decomposition approach we advocate, this problem can be further simplified: we only require a computationally efficient algorithm to map an input weighted sum of vectors through the non-linearity to an output weighted sum of vectors.\footnote{
    One could simply treat the effect of a non-linear activation as if it were an offset. 
    For instance, in the case of ReLU:
    \begin{equation*}
        \mathrm{ReLU} \left(\mathbf{v}\right) = \mathbf{v} + \mathbf{z}  \quad \text{where} ~ \mathbf{z}  = \mathrm{ReLU} \left(\mathbf{v}\right) - \mathbf{v} = \mathrm{ReLU}(-\mathbf{v})
    \end{equation*}
}

Also remark that previous research stressed that Transformer layers exhibit a certain degree of commutativity \citep{9533563} and that additional computation can be injected between contiguous sublayers \citep{pfeiffer2020AdapterHub}.
This can be thought of as evidence pointing towards a certain independence of the computations done in each layers: if we can shuffle and add layers, then it seems reasonable to characterize sub-layers based on what their outputs add to the total embedding, as we do in \Cref{eq:global}.

Beyond the expectations we may have, it remains to be seen whether our proposed methodology is of actual use, i.e., whether is conducive to further research. 
The remainder of this article presents some analyses that our decomposition enables us to conduct.\footnote{
    Code for our experiments is available at the following URL: \url{https://github.com/TimotheeMickus/bert-splat}.
}

\section{Visualizing the contents of embeddings}  \label{sec:exp-viz}

One major question is that of the relative relevance of the different submodules of the architecture with respect to the overall output embedding. 
Studying the four terms $\mathbf{i}_t$, $\mathbf{f}_t$, $\mathbf{h}_t$ and $\mathbf{c}_t$ can prove helpful in this endeavor. 
Given that \Cref{eq:i-term,eq:f-term,eq:h-term,eq:c-term} are defined as sums across layers or sublayers, it is straightforward to adapt them to derive the decomposition for intermediate representations.
Hence, we can study how relevant are each of the four terms to intermediary representations, and plot how this relevance evolves across layers.

To that end, we propose an \emph{importance metric} to compare one of the terms $\mathbf{t}_t$ to the total $\mathbf{e}_t$.
We require it to be sensitive to co-directionality (i.e., whether $\mathbf{t}_t$ and $\mathbf{e}_t$ have similar directions) and relative magnitude (whether $\mathbf{t}_t$ is a major component of $\mathbf{e}_t$).
%
A normalized dot-product of the form:
\begin{equation} \label{eq:norm-dot-prod}
    \mu(\mathbf{e}_t, \mathbf{t}_t) = {\mathbf{e}_t^T\mathbf{t}_t^{}} / {\lVert \mathbf{e}_t^{} \rVert^2_2}
\end{equation}
satisfies both of these requirements.
As dot-product distributes over addition (i.e., $\mathbf{a}^T \sum_i \mathbf{b}_i = \sum_i  \mathbf{a}^T \mathbf{b}_i$) and the dot-product of a vector with itself is its magnitude squared (i.e., $\mathbf{a}^T  \mathbf{a} = \lVert \mathbf{a} \rVert^2_2$): 
\begin{equation*}
    \mu(\mathbf{e}_t, \mathbf{i}_t) + \mu(\mathbf{e}_t, \mathbf{f}_t) + \mu(\mathbf{e}_t, \mathbf{h}_t) + \mu(\mathbf{e}_t, \mathbf{c}_t) = 1
\end{equation*}
Hence this function intuitively measures the importance of a term relative to the total.

We use the same Europarl sample as in \Cref{subsec:limits}.
We contrast embeddings from three related models: the BERT base uncased model and fine-tuned variants on CONLL 2003 NER \citep{tjong-kim-sang-de-meulder-2003-introduction}\footnote{\url{https://huggingface.co/dslim/bert-base-NER-uncased}} and SQuAD v2 \citep{rajpurkar-etal-2018-know}.\footnote{\url{https://huggingface.co/twmkn9/bert-base-uncased-squad2}}

\begin{table*}[thbp]
    \centering
    \nprounddigits{2}
\npdecimalsign{.}
\resizebox{\textwidth}{!}{
    \begin{tabular}{l l n{2}{2} n{2}{2} n{2}{2} n{2}{2} n{2}{2} n{2}{2} n{2}{2} n{2}{2} n{2}{2} n{2}{2} n{2}{2} n{2}{2} n{2}{2} n{2}{2} n{2}{2}}
    \multirow{4}{*}{\rotatebox{90}{Using}} 
    & {{\rotatebox{90}{$\mathbf{i}_t$}}} & \cmark & \xmark & \xmark & \xmark & \cmark & \cmark & \cmark & \xmark & \xmark & \xmark & \cmark & \cmark & \cmark & \xmark & \cmark \\
    & {{\rotatebox{90}{$\mathbf{h}_t$}}} & \xmark & \cmark & \xmark & \xmark & \cmark & \xmark & \xmark & \cmark & \cmark & \xmark & \cmark & \cmark & \xmark & \cmark & \cmark \\
    & {{\rotatebox{90}{$\mathbf{f}_t$}}} & \xmark & \xmark & \cmark & \xmark & \xmark & \cmark & \xmark & \cmark & \xmark & \cmark & \cmark & \xmark & \cmark & \cmark & \cmark \\
    & {{\rotatebox{90}{$\mathbf{c}_t$}}} & \xmark & \xmark & \xmark & \cmark & \xmark & \xmark & \cmark & \xmark & \cmark & \cmark & \xmark & \cmark & \cmark & \cmark & \cmark \\
\toprule
 \multicolumn{2}{l}{Default} &
{\gradC} 3.0847723994936262 & {\gradJ} 39.638012647628784 & {\gradA} 1.3610099176211017 & {\gradE} 4.444614851049015 & {\gradK} 40.554130928856985 & {\gradB} 2.9882542522890226 & {\gradD} 4.4046981526272636 & {\gradF} 22.19710839646203 & {\gradI} 35.688625063214985 & {\gradL} 48.77682072775705 & {\gradG} 32.450451595442636 & {\gradH} 35.233654720442636 & {\gradM} 49.408483505249023 & {\gradN} 57.16843008995056 & {\gradO} 57.33003275735038 \\
\midrule

    \multirow{7}{*}{\rotatebox{90}{Learned}} & \multirow{5}{*}{\rotatebox{90}{runs}} & 
{\gradC} 14.51466040951865 & {\gradE} 48.328796880585806 & {\gradI} 53.924389396395 & {\gradA} 9.239169423069273 & {\gradG} 49.715772696903776 & {\gradH} 53.66473197937012 & {\gradB} 13.184626400470734 & {\gradL} 55.45087456703186 & {\gradD} 48.09900607381548 & {\gradJ\bestCell{53.96701097488403}} & {\gradN} 55.58030435017177 & {\gradF} 49.08254402024405 & {\gradK} 54.05265263148716 & {\gradO\bestCell{55.79230189323425}} & {\gradM} 55.52683557782855 \\
    & & {\gradC\bestCell{15.5377453991345}} & {\gradG\bestCell{49.74250708307539}} & {\gradI} 53.73721974236625  & {\gradA} 9.629424874271665  & {\gradD} 47.7885548557554  & {\gradJ\bestCell{53.95616378102984}} & {\gradB} 13.560531820569718  & {\gradM} 55.52993927683149  & {\gradE} 48.92094135284424 & {\gradH} 53.41828550611224  & {\gradN} 55.58224064963204  & {\gradF} 49.43208524159023 & {\gradK} 54.02902109282357  & {\gradL} 55.32878977911813 & {\gradO\bestCell{55.90544598443168}} \\
    & & {\gradC} 15.08781441620418  & {\gradF} 49.60339665412903  & {\gradJ} 53.69263376508441  & {\gradA}  9.858846025807518  & {\gradG} 49.70762985093253 & {\gradI} 53.6108638559069 & {\gradB} 13.71826252767018 & {\gradN} 55.53885102272034  & {\gradE} 48.66595353399004 & {\gradH} 53.56941052845546  & {\gradL} 54.75643362317767 & {\gradD} 48.457828164100647  & {\gradK} 54.06740137508937 & {\gradM} 55.26370491300311 & {\gradO} 55.74811441557748 \\
&& {\gradC} 14.953344847474778 & {\gradD} 49.233299493789673 & {\gradH} 53.5577654838562 & {\gradA} 9.661967200892312 & {\gradE} 49.78435976164682 & {\gradJ} 53.85773863111224 & {\gradB} 11.521736213139125 & {\gradM\bestCell{55.60937353542873}} & {\gradG\bestCell{49.82894573892866}} & {\gradI} 53.65505218505859 & {\gradN} 55.72251762662616 & {\gradF} 49.827778339385986 & {\gradK} 53.90423195702689 & {\gradL} 55.48498289925712 & {\gradO} 55.72952287537711 \\
&& {\gradC}  15.218810737133026 & {\gradD} 49.51658844947815 & {\gradJ\bestCell{53.95229118210929}} & {\gradA\bestCell{10.207248798438481}} & {\gradF\bestCell{49.857616424560547}} & {\gradI} 53.85153208460126 & {\gradB\bestCell{14.2910469855581}} & {\gradL} 55.40204644203186 & {\gradE}  49.681663513183594 & {\gradH} 53.85036468505859 & {\gradO\bestCell{55.81049408231463}} & {\gradG\bestCell{49.862257071903776}} & {\gradK\bestCell{54.08482466425214}} & {\gradN} 55.78726189477103 & {\gradM} 55.52373273032052
    \\
    \cmidrule{2-17}
    & $\mu$ &  {\gradC}  15.062475161893028 &  {\gradE}  49.28491771221161 & {\gradI} 53.77285991396222 & {\gradA} 9.71933126449585 &  {\gradG} 49.37078671795981 & {\gradJ} 53.78820606640407 &  {\gradB}  13.25524078948157 & {\gradM}  55.50621696880886 &  {\gradD}  49.0393020425524 &  {\gradH}  53.69202477591379 & {\gradL}  55.49039806638446 &  {\gradF}  49.33249856744494 & {\gradK} 54.02762634413583 & {\gradN}  55.53140827587673 & {\gradO}  55.68673031670708
    \\
    & $\sigma$ & {{$\pm$\numprint{0.33555181917142396}}} & {{$\pm$\numprint{0.5062251839685266}}} & {{$\pm$\numprint{0.14773546269802693}}} & {{$\pm$\numprint{0.31605603215598543}}} & {{$\pm$\numprint{0.7929668484075653}}} & {{$\pm$\numprint{0.12942246005296623}}} & {{$\pm$\numprint{0.9370710461500956}}} & {{$\pm$\numprint{0.07236475204526305}}} & {{$\pm$\numprint{0.6440062572234935}}} & {{$\pm$\numprint{0.19597109503081803}}} & {{$\pm$\numprint{0.37724341041694814}}} & {{$\pm$\numprint{0.5221856415798175}}} & {{$\pm$\numprint{0.06435547539093732}}} & {{$\pm$\numprint{0.22289065265104044}}} & {{$\pm$\numprint{0.14530677525067767}}} \\
\bottomrule
\end{tabular}}
    \caption{Masked language model accuracy (in \%). Cells in \underline{\textbf{underlined bold}} font indicate best performance per setup across runs. Cell color indicates the ranking of setups within a run. Rows marked $\mu$ contain average performance; rows marked $\sigma$ contains the standard deviation across runs.}
    \label{tab:mlm-acc}
\end{table*}

\Cref{fig:breakdown} summarizes the relative importance of the four terms of \cref{eq:global}, as measured by the normalized dot-product defined in \cref{eq:norm-dot-prod}; ticks on the x-axis correspond to different layers.
\Cref{fig:breakdown:base,fig:breakdown:ner,fig:breakdown:squad} display the evolution of our proportion metric across layers for all three BERT models, whereas \Cref{fig:correl:base-ner,fig:correl:base-squad,fig:correl:ner-squad} display how our normalized dot-product measurements correlate across pairs of models using Spearman's $\rho$.\footnote{
    Layer 0 is the layer normalization conducted before the first sublayer, hence $\mathbf{f}_t$ and $\mathbf{h}_t$ are undefined here.
}

Looking at \Cref{fig:breakdown:base}, we can make a few important observations.
The input term $\mathbf{i}_t$, which corresponds to a static embedding, initially dominates the full output, but quickly decreases in prominence, until it reaches $0.045$ at the last layer.
This should explain why lower-layers of Transformers generally give better performances on static word-type tasks \citep[a.o.]{vulic-etal-2020-probing}.
The $\mathbf{h}_t$ term is not as prominent as one could expect from the vast literature that focuses on MHA.
Its normalized dot-product is barely above what we observe for $\mathbf{c}_t$, and never averages above 0.3 across any layer.
This can be partly pinned down on the prominence of $\mathbf{f}_t$ and its normalized dot-product of 0.4 or above across most layers. 
As FF submodules are always the last component added to each hidden state, the sub-terms of $\mathbf{f}_t$ go through fewer LNs than those of $\mathbf{h}_t$, and thus undergo fewer scalar multiplications---which likely affects their magnitude.
Lastly, the term $\mathbf{c}_t$ is far from negligible: at layer 11, it is the most prominent term, and in the output embedding it makes up for up to 23\%.
Note that $\mathbf{c}_t$ defines a set of offsets embedded in a $2\Lambda$-dimensional hyperplane (cf. Appendix~\ref{sec:appendix-hyperplane}).
In BERT base, 23\% of the output can be expressed using a 50-dimensional vector, or 6.5\% of the 768 dimensions of the model.
This likely induces part of the anisotropy of Transformer embeddings \citep[e.g.]{ethayarajh-2019-contextual,timkey-van-schijndel-2021-bark}{\color{black}, as the $\mathbf{c}_t$ term pushes the embedding towards a specific region of the space.}

The fine-tuned models in \Cref{fig:breakdown:ner,fig:breakdown:squad} are found to impart a much lower proportion of the contextual embeddings to the $\mathbf{i}_t$ and $\mathbf{c}_t$ terms.
While $\mathbf{f}_t$ seems to dominate in the final embedding, looking at the correlations in \Cref{fig:correl:base-ner,fig:correl:base-squad} suggest that the $\mathbf{h}_t$ terms are those that undergo the most modifications.
Proportions assigned to the terms correlate with those assigned in the non-finetuned model more in the case of lower layers than higher layers (\Cref{fig:correl:base-ner,fig:correl:base-squad}). 
The required adaptations seem task-specific as the two fine-tuned models do not correlate highly with each other (\Cref{fig:correl:ner-squad}).
Lastly, updates in the NER model impact mostly layer 8 and upwards (\Cref{fig:correl:base-ner}),  whereas the QA model (\Cref{fig:correl:base-squad}) sees important modifications to the $\mathbf{h}_t$ term at the first layer, suggesting that SQuAD requires more drastic adaptations than CONLL 2003.




\section{The MLM objective}  \label{sec:exp-mlm}

An interesting follow-up question concerns 
%
which of the four terms allow us to retrieve the target word-piece.
We consider two approaches: (a) using the actual projection learned by the non-finetuned BERT model, or (b) learning a simple categorical regression for a specific term.
We randomly select 15\% of the word-pieces in our Europarl sample.
As in the work of \citet{devlin-etal-2019-bert}, 80\% of these items are masked, 10\% are replaced by a random word-piece and 10\% are left as is.
Selected embeddings are then split between train (80\%), validation (10\%) and test (10\%).

Results are displayed in \Cref{tab:mlm-acc}.
The first row (``Default'') details predictions using the default output projection on the vocabulary, i.e., we test the performances of combinations sub-terms under the circumstances encountered by the model during training.\footnote{
    We thank an anonymous reviewer for pointing out that the BERT model ties input and output embeddings; we leave investing the implications of this fact for future work.
}
The rows below (``Learned'') correspond to learned linear projections; the row marked $\mu$ display the average performance across all 5 runs.
Columns display the results of using the sum of 1, 2, 3 or 4 of the terms $\mathbf{i}_t$, $\mathbf{h}_t$, $\mathbf{f}_t$ and $\mathbf{c}_t$ to derive representations; e.g. the rightmost corresponds to $\mathbf{i}_t+\mathbf{h}_t+\mathbf{f}_t+\mathbf{c}_t$ (i.e., the full embedding), whereas the leftmost corresponds to predicting based on $\mathbf{i}_t$ alone.
Focusing on the default projection first, we see that it benefits from a more extensive training: when using all four terms, it is almost 2\% more accurate than learning one from scratch.
On the other hand, learning a regression allows us to consider more specifically what can be retrieved from individual terms, as is apparent from the behavior of the $\mathbf{f}_t$: when using the default output projection, we get 1.36\% accuracy, whereas a learned regression yields 53.77\%.

The default projection matrix is also highly dependent on the normalization offsets $\mathbf{c}_t$ and the FF terms $\mathbf{f}_t$ being added together: removing this $\mathbf{c}_t$ term from any experiment using $\mathbf{f}_t$ is highly detrimental to the accuracy.
On the other hand, combining the two produces the highest accuracy scores.
Our logistic regressions show that most of this performance can be imputed to the $\mathbf{f}_t$ term. Learning a projection from the $\mathbf{f}_t$ term already yields an accuracy of almost 54\%. 
On the other hand, a regression learned from $\mathbf{c}_t$ only has a limited performance of 9.72\% on average. 
Interestingly, this is still above what one would observe if the model always predicted the most frequent word-piece (viz. \texttt{the}, 6\% of the test targets): even these very semantically bare items can be exploited by a classifier.
As $\mathbf{c}_t$ is tied to the LN $z$-scaling, this suggests that the magnitude of Transformer embeddings is not wholly meaningless.

In all, do FFs make the model more effective?
The $\mathbf{f}_t$ term is necessary to achieve the highest accuracy on the training objective of BERT.
On its own, it doesn't achieve the highest performances: for that we also need to add the MHA outputs $\mathbf{h}_t$.
However, the performances we can associate to $\mathbf{f}_t$ on its own are higher than what we observe for $\mathbf{h}_t$, suggesting that FFs make the Transformer architecture more effective on the MLM objective.
This result connects with the work of \citet{geva-etal-2021-transformer,geva-etal-2022-transformer} who argue that FFs update the distribution over the full vocabulary, hence it makes sense that $\mathbf{f}_t$ would be most useful to the MLM task.


\begin{table*}[t]
    \centering
    \nprounddigits{2}
\npdecimalsign{.}
\resizebox{\textwidth}{!}{
    \begin{tabular}{l l n{2}{2} n{2}{2} n{2}{2} n{2}{2} n{2}{2} n{2}{2} n{2}{2} n{2}{2} n{2}{2} n{2}{2} n{2}{2} n{2}{2} n{2}{2} n{2}{2} n{2}{2}}
    \multirow{4}{*}{\rotatebox{90}{Using}} 
    & {{\rotatebox{90}{$\mathbf{i}_t$}}} & \cmark & \xmark & \xmark & \xmark & \cmark & \cmark & \cmark & \xmark & \xmark & \xmark & \cmark & \cmark & \cmark & \xmark & \cmark \\
    & {{\rotatebox{90}{$\mathbf{h}_t$}}} & \xmark & \cmark & \xmark & \xmark & \cmark & \xmark & \xmark & \cmark & \cmark & \xmark & \cmark & \cmark & \xmark & \cmark & \cmark \\
    & {{\rotatebox{90}{$\mathbf{f}_t$}}} & \xmark & \xmark & \cmark & \xmark & \xmark & \cmark & \xmark & \cmark & \xmark & \cmark & \cmark & \xmark & \cmark & \cmark & \cmark \\
    & {{\rotatebox{90}{$\mathbf{c}_t$}}} & \xmark & \xmark & \xmark & \cmark & \xmark & \xmark & \cmark & \xmark & \cmark & \cmark & \xmark & \cmark & \cmark & \cmark & \cmark \\
    \toprule
             \multicolumn{2}{l}{\textsc{knn}} & 
{\gradA} 54.36407576013039 & 
{\gradK} 64.07013994267409 & 
{\gradG} 62.45152588096442 & 
{\gradB} 55.398190299556 & 
{\gradM} 64.22188501095937 & 
{\gradE} 62.22109818467937 & 
{\gradC} 56.34238183555331 & 
{\gradH} 63.37323666610465 & 
{\gradN} 64.39611083010173 & 
{\gradF} 62.43466531782161 & 
{\gradI} 63.55870286067554 & 
{\gradO} 64.43545214410161 & 
{\gradD} 62.17613668296521 & 
{\gradL} 64.09824088124544 & 
{\gradJ} 63.93525543753161 \\
        \midrule

\multirow{7}{*}{\rotatebox{90}{Classifiers}} & \multirow{5}{*}{\rotatebox{90}{runs}} &
{\gradC\bestCell{58.415119735292464}} & {\gradN} 66.83865178621502 & {\gradF} 64.46077056811171 & {\gradA} 57.32151870338175 & {\gradL} 66.65358084235321 & {\gradD} 63.883124894846055 & {\gradB\bestCell{57.82625764118669}} & {\gradH} 65.71140149178397 & {\gradO\bestCell{66.94520778419606}} & {\gradG} 64.46077056811171 & {\gradI} 65.75626717514442 & {\gradM\bestCell{66.77135326117435}} & {\gradE} 64.4495541472716 & {\gradJ} 65.87964780438563 & {\gradK} 65.98620380236666
\\
 &&
{\gradC} 57.91038079748753 & {\gradL} 66.5189837922719 & {\gradG} 64.73557287869441 & {\gradA} 57.40003364926252 & {\gradM} 66.68723010487354 & {\gradD} 64.23083394088947 & {\gradB} 57.65801132858505 & {\gradI} 65.96377096068645 & {\gradO} 66.8050025236947 & {\gradE} 64.387863832651 & {\gradH\bestCell{65.93572990858617}} & {\gradN} 66.72648757781391 & {\gradF} 64.56171835567271 & {\gradJ} 65.98620380236666 & {\gradK} 66.02546127530705
\\
 &&
{\gradC} 58.308563737311424 & {\gradL} 66.38999495261062 & {\gradG} 64.6738825640738 & {\gradA\bestCell{57.405641859682575}} & {\gradN} 66.56384947563232 & {\gradF\bestCell{64.58975940777299}} & {\gradB} 57.54584712018394 & {\gradK\bestCell{66.13762548370815}} & {\gradM} 66.50776737143178 & {\gradE} 64.34860635971062 & {\gradJ} 65.89086422522573 & {\gradO} 66.73770399865403 & {\gradD} 64.16914362626885 & {\gradH} 65.7955246480848 & {\gradI} 65.83478212102519
\\
 &&
{\gradB} 57.73652627446582 & {\gradO\bestCell{67.05737199259715}} & {\gradE} 64.4383377264315 & {\gradA} 57.282261230441364 & {\gradL} 66.40681958387079 & {\gradD} 64.41590488475127 & {\gradC} 57.82625764118669 & {\gradK} 66.13201727328808 & {\gradN} 66.5189837922719 & {\gradG} 64.64023330155348 & {\gradJ} 65.88525601480568 & {\gradM} 66.45729347765128 & {\gradF\bestCell{64.58415119735292}} & {\gradI} 65.84039033144524 & {\gradH} 65.82917391060512
\\
 &&
{\gradC} 58.095451741349336 & {\gradL} 66.5918905277326 & {\gradG\bestCell{64.78043856205484}} & {\gradA} 57.19813807414054 & {\gradO\bestCell{67.09102125511748}} & {\gradD} 64.31495709719029 & {\gradB} 57.34395154506197 & {\gradI} 65.86282317312545 & {\gradN} 66.79939431327463 & {\gradF\bestCell{64.69631540575402}} & {\gradH} 65.73383433346419 & {\gradM} 66.73770399865403 & {\gradE} 64.57293477651281 & {\gradK\bestCell{66.12080085244799}} & {\gradJ\bestCell{66.08154337950761}}
\\
    \cmidrule{2-17}
    & $\mu$ & {\gradC} 58.09320845718131 & {\gradL} 66.67937861028545 & {\gradG} 64.61780045987325 & {\gradA} 57.321518703381756 & {\gradM} 66.68050025236947 & {\gradD} 64.28691604509001 & {\gradB} 57.64006505524087 & {\gradK} 65.96152767651841 & {\gradO} 66.71527115697381 & {\gradF} 64.50675789355617 & {\gradH} 65.84039033144524 & {\gradN} 66.68610846278952 & {\gradE} 64.46750042061578 & {\gradI} 65.92451348774605 & {\gradJ} 65.95143289776232 \\
    & $\sigma$ & {{$\pm$\numprint{0.24926208399783928}}} & {{$\pm$\numprint{0.23890186487584097}}} & {{$\pm$\numprint{0.14165589075559415}}} & {{$\pm$\numprint{0.07746641982987944}}} & {{$\pm$\numprint{0.227070892573597}}} & {{$\pm$\numprint{0.23466189256378156}}} & {{$\pm$\numprint{0.18232808727466568}}} & {{$\pm$\numprint{0.16271930426807737}}} & {{$\pm$\numprint{0.1729657923948243}}} & {{$\pm$\numprint{0.13784794118770383}}} & {{$\pm$\numprint{0.08010126665406082}}} & {{$\pm$\numprint{0.1153929327540729}}} & {{$\pm$\numprint{0.15680540288084058}}} & {{$\pm$\numprint{0.11667234478331771}}} & {{$\pm$\numprint{0.10214951551816707}}} \\
\bottomrule
\end{tabular}}
    \caption{Accuracy on SemCor WSD (in \%)}
    \label{tab:wsd-acc}
\end{table*}

\section{Lexical contents \& WSD}  \label{sec:exp-wsd}


We now turn to look at how the vector spaces are organized, and which term yields the most linguistically appropriate space.
We rely on \textsc{wsd}, as distinct senses should yield different representations.

We consider an intrinsic KNN-based setup and an extrinsic probe-based setup. 
The former is inspired from \citet{wiedemann-2019-bert}: we assign to a target the most common label in its neighborhood.
We restrict neighborhoods to words with the same annotated lemma and use the 5 nearest neighbors using cosine distance.
The latter is a 2-layer MLP similar to \citet{du-etal-2019-using}, where the first layer is shared for all items and the second layer is lemma-specific. 
We use the \textsc{nltk} Semcor dataset \citep{landes-etal-1998-building,bird-etal-2009-natural}, with an 80\%--10\%--10\% split. 
We drop monosemous or OOV lemmas and sum over word-pieces to convert them into single word representations.
\Cref{tab:wsd-acc} shows accuracy results.
Selecting the most frequent sense would yield an accuracy of 57\%; picking a sense at random, 24\%.
The terms $\mathbf{i}_t$ and $\mathbf{c}_t$ struggle to outperform the former baseline: relevant KNN accuracy scores are lower, and corresponding probe accuracy scores are barely above.
 
Overall the same picture emerges from the {\color{black}KNN} setup and all 5 runs of the {\color{black}classifier} setup.
The $\mathbf{f}_t$ term does not yield the highest performances in our experiment, instead, the $\mathbf{h}_t$ term systematically dominates.
In single term models, $\mathbf{h}_t$ is ranked first and $\mathbf{f}_t$ second. 
As for sums of two terms, the setups ranked 1\textsuperscript{st}, 2\textsuperscript{nd}, and 3\textsuperscript{rd} are those that include $\mathbf{h}_t$; setups ranked 3\textsuperscript{rd} to 5\textsuperscript{th}, those that include $\mathbf{f}_t$.
Even more surprisingly, when summing three of the terms, the highest ranked setup is the one where we exclude $\mathbf{f}_t$, and the lowest corresponds to excluding $\mathbf{h}_t$.
Removing {\color{black}$\mathbf{f}_t$} systematically yields better performances than using the full embedding
This suggests that $\mathbf{f}_t$ is not necessarily helpful to the final representation for WSD.
{\color{black} This contrast with what we observed for MLM, where $\mathbf{h}_t$ was found to be less useful then $\mathbf{f}_t$.}

\begin{figure}
    \centering
    %
    \resizebox{\columnwidth}{!}{\input{figs/heatmaps/heatmap-semcor-pgf}}
    \caption{Prediction agreement for WSD models (in \%). Upper triangle: agreement for KNNs; lower triangle: for learned classifiers.}
    \label{fig:wsd:pred-agree}
\end{figure}

One argument that could be made here would be to posit that the predictions derived from the different sums of terms are intrinsically different, hence a purely quantitative ranking might not capture this important distinction. 
To verify whether this holds, we can look at the proportion of predictions that agree for any two models.
Since our intent is to see what can be retrieved from specific subterms of the embedding, we focus solely on the most efficient classifiers across runs.
This is summarized in \Cref{fig:wsd:pred-agree}: an individual cell will detail the proportion of the assigned labels shared by the models for that row and that column.
In short, we see that model predictions tend to a high degree of overlap. 
For both \textsc{knn} and classifier setups, the three models which appear to make the most distinct predictions turn out to be computed from the $\mathbf{i}_t$ term, the $\mathbf{c}_t$ term or their sum: i.e., the models that struggle to perform better than the MFS baseline and are derived from static representations. 

\section{Effects of finetuning \& NER}  \label{sec:exp-ner}

Downstream application can also be achieved through fine-tuning, i.e., restarting a model's training to derive better predictions on a narrower task.
As we saw from \Cref{fig:breakdown:ner,fig:breakdown:squad}, the modifications brought upon this second round of training are task specific, meaning that an exhaustive experimental survey is out of our reach.

We consider the task of Named Entity Recognition, using the WNUT 2016 shared task dataset \citep{strauss-etal-2016-results}.
We contrast the performances of the non-finetuned BERT model to that of the aforementioned variant finetuned on the CONLL 2003 NER dataset using shallow probes. 

\begin{table*}[t]
    \centering
        \nprounddigits{2}
    \npdecimalsign{.}
    \nprounddigits{2}
    \npdecimalsign{.}
\resizebox{\textwidth}{!}{
    \begin{tabular}{l l n{2}{2} n{2}{2} n{2}{2} n{2}{2} n{2}{2} n{2}{2} n{2}{2} n{2}{2} n{2}{2} n{2}{2} n{2}{2} n{2}{2} n{2}{2} n{2}{2} n{2}{2}}
    \multirow{4}{*}{\rotatebox{90}{Using}} 
    & {{\rotatebox{90}{$\mathbf{i}_t$}}} & \cmark & \xmark & \xmark & \xmark & \cmark & \cmark & \cmark & \xmark & \xmark & \xmark & \cmark & \cmark & \cmark & \xmark & \cmark \\
    & {{\rotatebox{90}{$\mathbf{h}_t$}}} & \xmark & \cmark & \xmark & \xmark & \cmark & \xmark & \xmark & \cmark & \cmark & \xmark & \cmark & \cmark & \xmark & \cmark & \cmark \\
    & {{\rotatebox{90}{$\mathbf{f}_t$}}} & \xmark & \xmark & \cmark & \xmark & \xmark & \cmark & \xmark & \cmark & \xmark & \cmark & \cmark & \xmark & \cmark & \cmark & \cmark \\
    & {{\rotatebox{90}{$\mathbf{c}_t$}}} & \xmark & \xmark & \xmark & \cmark & \xmark & \xmark & \cmark & \xmark & \cmark & \cmark & \xmark & \cmark & \cmark & \cmark & \cmark \\
    \toprule
    
\multirow{7}{*}{\rotatebox{90}{Base}} & \multirow{5}{*}{\rotatebox{90}{runs}} &
{\gradC} 14.925487293085993 & {\gradK} 32.89274940603739 & {\gradF} 31.11541102910062 & {\gradA} 4.924466891933662 & {\gradD} 20.954431035502612 & {\gradE} 30.623264378747727 & {\gradB} 12.806387109056699 & {\gradH} 31.768447168677316 & {\gradO\bestCell{34.62474424197677}} & {\gradI\bestCell{32.42514452066802}} & {\gradN} 34.0286710283412 & {\gradJ} 32.52460058284885 & {\gradG} 31.44503037906211 & {\gradM} 33.8473243861131 & {\gradL} 32.947501604670684 \\
 && 
{\gradC\bestCell{15.790921559066351}} & {\gradM\bestCell{34.156410085837294}} & {\gradG} 31.05029159515828 & {\gradA} 4.8796685520470815 & {\gradO\bestCell{34.384148690759126}} & {\gradH} 31.07969823890693 & {\gradB} 10.261161846763434 & {\gradN} 34.308950588092124 & {\gradK} 33.12237671365441 & {\gradD} 30.13882118146335 & {\gradI} 31.844803043188307 & {\gradJ} 32.57258354841511 & {\gradE} 30.48107932510228 & {\gradL\bestCell{33.99345960747829}} & {\gradF} 30.819491947859873 \\
 && 
{\gradC} 15.458678510110394 & {\gradF} 29.94263956799924 & {\gradE} 28.9205555410673 & {\gradA} 4.917086614618284 & {\gradM} 33.728051946940376 & {\gradG} 30.488305930170483 & {\gradB\bestCell{13.71941860539126}} & {\gradH} 31.47998387678794 & {\gradI} 32.18006277225617 & {\gradD} 17.350763048165785 & {\gradO\bestCell{34.67656003045336}} & {\gradN\bestCell{34.46144009497011}} & {\gradK\bestCell{32.71835818228324}} & {\gradL} 32.82305186612183 & {\gradJ} 32.41781934495527 \\
 && 
{\gradC} 15.549833563768962 & {\gradM} 33.95086724746083 & {\gradE} 31.358989787669827 & {\gradA\ \ \bestCell{5.324272164802305}} & {\gradK} 33.16013653829686 & {\gradI\bestCell{32.03187423520723}} & {\gradB} 12.884591584949442 & {\gradO\bestCell{34.317336543551754}} & {\gradL} 33.37172396518168 & {\gradD} 31.08565947345796 & {\gradH} 31.802510045449434 & {\gradJ} 32.9099549108776 & {\gradF} 31.423155955813954 & {\gradG} 31.720035956875535 & {\gradN\bestCell{34.00727397916435}} \\
 && 
{\gradA} 4.5206564799094515 & {\gradM} 33.6245710522088 & {\gradJ\bestCell{31.375827359664086}} & {\gradB} 5.281808160854706 & {\gradF} 28.222159034535032 & {\gradE} 27.997769982633276 & {\gradC} 12.198530314513148 & {\gradK} 31.994845730426917 & {\gradN} 33.69700842848452 & {\gradG} 30.44969953592391 & {\gradH} 31.19183889009065 & {\gradO} 33.80190766962302 & {\gradL} 32.71127477054909 & {\gradD} 13.62734046649534 & {\gradI} 31.326034275546128 \\

\cmidrule{2-17} 
& $\mu$ & {\gradC} 13.249115481188229 & {\gradM} 32.91344747190871 & {\gradH} 30.764215062532024 & {\gradA} 5.065460476851206 & {\gradF} 30.0897854492068 & {\gradG} 30.444182553133132 & {\gradB} 12.374017892134797 & {\gradL} 32.773912781507214 & {\gradO} 33.39918322431071 & {\gradD} 28.290017551935804 & {\gradK} 32.70887660750459 & {\gradN} 33.25409736134694 & {\gradI} 31.75577972256213 & {\gradE} 29.202242456616816 & {\gradJ} 32.30362423043926
 \\
& $\sigma$ & {{$\pm$\numprint{4.373384721255649}}} & {{$\pm$\numprint{1.5461578579596194}}} & {{$\pm$\numprint{0.9308152766480522}}} & {{$\pm$\numprint{0.19503971057095618}}} & {{$\pm$\numprint{5.060599238389999}}} & {{$\pm$\numprint{1.3374301254239922}}} & {{$\pm$\numprint{1.1621263505427342}}} & {{$\pm$\numprint{1.2673323601975217}}} & {{$\pm$\numprint{0.7942707588895789}}} & {{$\pm$\numprint{5.525657952347001}}} & {{$\pm$\numprint{1.3771827962646972}}} & {{$\pm$\numprint{0.7580017498076391}}} & {{$\pm$\numprint{0.856923906286014}}} & {{$\pm$\numprint{7.8301936733421496}}} & {{$\pm$\numprint{1.1391551093384786}}}
\\ 
\midrule
\multirow{7}{*}{\rotatebox{90}{Tuned}} & \multirow{5}{*}{\rotatebox{90}{runs}} &
{\gradC\bestCell{16.27753618335451}} & {\gradM} 39.23859083686958 & {\gradD} 36.962535752528076 & {\gradA} 12.229267478987357 & {\gradN\bestCell{39.922041727698556}} & {\gradI} 38.65491696901654 & {\gradB} 15.698953548896993 & {\gradF} 37.39018182789866 & {\gradK} 38.673786447637944 & {\gradH} 38.2483696626909 & {\gradJ} 38.669882762078565 & {\gradO\bestCell{40.173118405709715}} & {\gradL} 39.00585098563675 & {\gradE} 37.29527147504887 & {\gradG} 37.45978676674074
\\
 &&
{\gradA} 5.065561627975216 & {\gradO} 39.62258914360729 & {\gradK} 38.80035528309326 & {\gradC} 8.689602400372053 & {\gradH} 38.58216490006029 & {\gradN\bestCell{39.496565904296304}} & {\gradB} 6.356744201261226 & {\gradE} 37.673636551208176 & {\gradL} 38.86235118804534 & {\gradF} 37.851658262985914 & {\gradJ} 38.767700508415324 & {\gradG} 38.01439853098153 & {\gradM} 39.30704365540667 & {\gradD} 37.65002298830424 & {\gradI} 38.59927839956613
\\
 &&
{\gradC} 16.062039281524264 & {\gradK} 39.05720218355015 & {\gradI} 38.61676179911783 & {\gradB} 11.80910605528667 & {\gradJ} 38.925735725447325 & {\gradF} 37.52643442470734 & {\gradA} 4.685581363559768 & {\gradO\bestCell{39.91779448965958}} & {\gradN\bestCell{39.77515788295233}} & {\gradG} 37.91846829171353 & {\gradH} 38.028941837882954 & {\gradM} 39.32573626181324 & {\gradD} 35.825093596993376 & {\gradE} 37.17782706093058 & {\gradL\bestCell{39.26987836303897}}
\\
 &&
{\gradB} 15.081061795434575 & {\gradH} 38.24272563231973 & {\gradN\bestCell{39.809098547559294}} & {\gradA} 4.826073349799207 & {\gradM} 39.75906401529388 & {\gradJ} 38.73593696083093 & {\gradC\bestCell{16.07421274942172}} & {\gradG} 37.92121314440826 & {\gradE} 37.22734057796887 & {\gradO\bestCell{39.94433834978815}} & {\gradF} 37.49077446148982 & {\gradL} 38.99933715593386 & {\gradI} 38.427124158127675 & {\gradD} 36.36895871621794 & {\gradK} 38.97628654999753
\\
 &&
{\gradA} 11.675033215641204 & {\gradO\bestCell{40.24768727443249}} & {\gradL} 39.23213672108899 & {\gradB\bestCell{12.360450538869582}} & {\gradJ} 38.87211229858171 & {\gradE} 37.43293558234941 & {\gradC} 15.744816948402834 & {\gradD} 35.48349878897752 & {\gradM} 39.31936933656737 & {\gradI} 38.827777931260556 & {\gradK\bestCell{38.889215775025534}} & {\gradH} 38.80471103669675 & {\gradN\bestCell{39.470229302539764}} & {\gradF\bestCell{38.26205563747604}} & {\gradG} 38.2989146504183
\\
\cmidrule{2-17} 
& $\mu$ &
{\gradC} 12.832246420785955 & {\gradO} 39.281759014155845 & {\gradK} 38.684177620677495 & {\gradA} 9.982899964662973 & {\gradN} 39.21222373341635 & {\gradG} 38.36935796824011 & {\gradB} 11.712061762308508 & {\gradE} 37.677264960430435 & {\gradL} 38.77160108663437 & {\gradJ} 38.55812249968781 & {\gradF} 38.369303068978446 & {\gradM} 39.06346027822702 & {\gradH} 38.407068339740846 & {\gradD} 37.35082717559553 & {\gradI} 38.52082894595234
 \\
& $\sigma$ & {{$\pm$\numprint{4.219461793490511}}} & {{$\pm$\numprint{0.6604638084580609}}} & {{$\pm$\numprint{0.9537084598415136}}} & {{$\pm$\numprint{2.9085880338515206}}} & {{$\pm$\numprint{0.5286967333351028}}} & {{$\pm$\numprint{0.7840827528877896}}} & {{$\pm$\numprint{5.084046316015852}}} & {{$\pm$\numprint{1.412410931220993}}} & {{$\pm$\numprint{0.8613617245040047}}} & {{$\pm$\numprint{0.7744035741642357}}} & {{$\pm$\numprint{0.5304783078264314}}} & {{$\pm$\numprint{0.7030964863791304}}} & {{$\pm$\numprint{1.3390894866804424}}} & {{$\pm$\numprint{0.6191827825676969}}} & {{$\pm$\numprint{0.6244561530998278}}}
\\ 
\bottomrule
    \end{tabular}}

    \caption{Macro-$f_1$ on WNUT 2016 (in \%)}
    \label{tab:ner-f1}
\end{table*}

Results are presented in \Cref{tab:ner-f1}.
The very high variance we observe across is likely due to the smaller size of this dataset (46,469 training examples, as compared to the 142,642 of \Cref{sec:exp-wsd} or the 107,815 in \Cref{sec:exp-mlm}).
Finetuning BERT on another NER dataset unsurprisingly has a systematic positive impact: average performance jumps up by 5\% or more.
More interesting is the impact this fine-tuning has on the $\mathbf{f}_t$ term: when used as sole input, the highest observed performance increases by over 8\%, and similar improvements are observed consistently across all setups involving $\mathbf{f}_t$.
Yet, the best average performance for fine-tuned and base embeddings correspond to $ \mathbf{h}_t$ (39.28\% in tuned), $\mathbf{i}_t+\mathbf{h}_t$ (39.21\%), and $\mathbf{i}_t+\mathbf{h}_t +\mathbf{c}_t$ (39.06\%); in the base setting the highest average performance are reached with $\mathbf{h}_t +\mathbf{c}_t$ (33.40\%), $\mathbf{i}_t + \mathbf{h}_t +\mathbf{c}_t$ (33.25\%) and $\mathbf{h}_t$ (32.91\%)---suggesting that $\mathbf{f}_t$ might be superfluous for this task.

\begin{figure}
    \centering
    
        \resizebox{\columnwidth}{!}{\input{figs/heatmaps/heatmap-wnut-pgf}}
    
    \caption{NER prediction agreement (macro-average, in \%). Upper triangle: agreement for untuned models; lower triangle: for tuned models.}
    \label{fig:ner:pred-agree}
\end{figure}

We can also look at whether the  highest scoring classifiers across runs classifiers produce different outputs. 
Given the high class imbalance of the dataset at hand, we macro-average the prediction overlaps by label.
The result is shown in \Cref{fig:ner:pred-agree}; 
the upper triangle details the behavior of the untuned model, whereas the lower triangle details that of the NER-finetuned model. 
In this round of experiments, we see much more distinctly that the $\mathbf{i}_t$ model, the $\mathbf{c}_t$ model and the $\mathbf{i}_t+\mathbf{c}_t$ model behave markedly different from the rest, with $\mathbf{c}_t$ yielding the most distinct predictions.
As for the NER-finetuned model (lower triangle), aside from the aforementioned static representations, most predictions display a degree of overlap much higher than what we observe for the non-finetuned model: both FFs and MHAs are skewed towards producing outputs more adapted to NER tasks.

\section{Relevant works} \label{sec:sota}
The derivation we provide in \Cref{sec:transformer} ties in well with other studies setting out to explain how Transformers embedding spaces are structured \citep[a.o.]{voita-etal-2019-bottom,mickus-etal-2020-mean,vazquez-etal-2021-differences} and more broadly how they behave \citep{rogers-etal-2020-primer}.
For instance, lower layers tend to yield higher performance on surface tasks (e.g., predicting the presence of a word, \citealt{jawahar-etal-2019-bert}) or static benchmarks (e.g., analogy, \citealt{vulic-etal-2020-probing}): this ties in with the vanishing prominence of $\mathbf{i}_t$ across layers.
Likewise, probe-based approaches to unearth a linear structure matching with the syntactic structure of the input sentence \citep[a.o.]{raganato-tiedemann-2018-analysis,hewitt-manning-2019-structural} can be construed as relying on the explicit linear dependence that we highlight here.

Another connection is with studies on embedding space anisotropy \citep{ethayarajh-2019-contextual,timkey-van-schijndel-2021-bark}: our derivation provides a means of circumscribing which neural components are likely to cause it.
Also relevant is the study on sparsifying Transformer representations of \citet{yun-etal-2021-transformer}: the linearly dependent nature of Transformer embeddings has some implications when it comes to dictionary coding.

Also relevant are the works focusing on the interpretation of specific Transformer components, and feed-forward sublayers in particular \citep{geva-etal-2021-transformer,9533563,geva-etal-2022-transformer}.
Lastly, our approach provides some quantitative argument for the validity of attention-based studies \citep{serrano-smith-2019-attention,jain-wallace-2019-attention,wiegreffe-pinter-2019-attention,pruthi-etal-2020-learning} and expands on earlier works looking beyond attention weights \citep{kobayashi-etal-2020-attention}.

\section{Conclusions and Future Work}  \label{sec:ccl}

In this paper, we stress how Transformer embeddings can be decomposed linearly to describe the impact of each network component.
We showcased how this additive structure can be used to investigate Transformers. 
Our approach suggests a less central place for attention-based studies: if multi-head attention only accounts for 30\% of embeddings, can we possibly explain what Transformers do by looking solely at these submodules?
The crux of our methodology lies in that we decompose the output embedding by submodule instead of layer or head.
These approaches are not mutually exclusive (cf. \Cref{sec:exp-viz}), hence our approach can easily be combined with other probing protocols, providing the means to narrow in on specific network components. 


The experiments we have conducted in \Cref{sec:exp-viz,sec:exp-mlm,sec:exp-wsd,sec:exp-ner} were designed so as to showcase whether our decomposition in \Cref{eq:global} could yield useful results---or, as we put it earlier in \Cref{subsec:limits}, whether this approach could be conducive to future research.
We were able to use the proposed approach to draw insightful connections.
The noticeable anisotropy of contextual embeddings can be connected to the prominent trace of the biases in the output embedding:
as model biases make up an important part of the whole embedding, they push it towards a specific sub-region of the embedding.
The diminishing importance of $\mathbf{i}_t$ links back to earlier results on word-type semantic benchmarks.
We also report novel findings, showcasing how some submodules outputs may be detrimental in specific scenarios: the output trace of FF modules was found to be extremely useful for MLM, whereas the $\mathbf{h}_t$ term was found to be crucial for WSD.
Our methodology also allows for an overview of the impact of finetuning (cf. \Cref{sec:exp-ner}): it skews components towards more task-specific outputs, and its effect are especially noticeable in upper layers (\Cref{fig:correl:base-ner,fig:correl:base-squad}).

Analyses in \Cref{sec:exp-viz,sec:exp-mlm,sec:exp-wsd,sec:exp-ner} demonstrate the immediate insight that our Transformer decomposition can help achieve. 
This work therefore opens a number of research perspectives, of which we name three.
First, as mentioned in \cref{subsec:limits}, our approach can be extended further to more thoroughly disentangle computations.
Second, while we focused here more on feed-forward and multi-head attention components, extracting the static component embeddings from $\mathbf{i}_t$ would allow for a principled comparison of contextual and static distributional semantics models.
Last but not least, since our analysis highlights the different relative importance of Transformer components in different tasks, it can be employed to help choose the most appropriate tools for further interpretation of trained models among the wealth of alternatives.

\section*{Acknowledgments}

We are highly indebted to Marianne Clausel for her significant help with how best to present the mathematical aspects of this work.
Our thanks also go to Aman Sinha, as well as three anonymous reviewers for their substantial comments towards bettering this work.

This work was supported  by a public grant overseen by the French National Research Agency (ANR) as part of the ``Investissements d'Avenir'' program: Idex \emph{Lorraine Universit\'e d'Excellence} (reference: {ANR-15-IDEX-0004}).
We also acknowledge the support by the FoTran project, funded by the European Research Council (ERC) under the European Union’s Horizon 2020 research and innovation programme (grant agreement n° 771113).

\bibliography{tacl2021}
\bibliographystyle{acl_natbib}

\appendix

\section{Step-by-step derivation of \cref{eq:global}}
\label{sec:appendix-deriv}
Given that a Transformer layer consists of a stack of $L$ layers, each comprised of two sublayers, we can treat a Transformer as a stack of $\Lambda=2L$ sublayers.
For notation simplicity, we link the sublayer index $\lambda$ to the layer index $l$: the first sublayer of layer $l$ is the $(2l-1)$\textsuperscript{th} sublayer, and the second is the $(2l)$\textsuperscript{th} sublayer.\footnote{
    In the case of BERT, we also need to include a LN before the first layer, which is straightforward if we index it as $\lambda=0$.
}
All sublayers include a residual connection before the final LN:
{\smaller\begin{equation*}
    \mathbf{y}_{\lambda,t} = \mathbf{g}_\lambda \odot \left[\frac{\left(S\left(\mathbf{x}\right) + \mathbf{x} \right) - m_{\lambda,t} \cdot \vec{1}}{s_{\lambda,t}}\right] + \mathbf{b}_\lambda^{(\text{LN})}
\end{equation*}}
We can model the effects of the gain $\mathbf{g}_\lambda$ and the scaling $1/s_{\lambda,t}$ as the $d \times d$ square matrix:
{\smaller\begin{equation*}
    \mathbf{T}_\lambda = \frac{1}{s_{\lambda,t}} \begin{bmatrix}
        \left(\mathbf{g}_\lambda\right)_1 & 0 & \hdots & 0 \\
        0  & \left(\mathbf{g}_\lambda\right)_2 & \\
        \vdots & & \ddots \\
        0 & & & \left(\mathbf{g}_\lambda\right)_d \\
    \end{bmatrix}
\end{equation*}}
which we use to rewrite a sublayer output $\mathbf{y}_{\lambda,t}$ as:
{\smaller\begin{align*} 
    \mathbf{y}_{\lambda,t} &= \left(S_\lambda\left(\mathbf{x}\right) + \mathbf{x} - \left(m_{\lambda,t} \cdot \vec{1} \right) \right)\mathbf{T}_\lambda + \mathbf{b}_\lambda^{(\text{LN})} \\
            &= S_\lambda\left(\mathbf{x}\right)\mathbf{T}_\lambda  +  \mathbf{x} \mathbf{T}_\lambda - \left(m_{\lambda,t} \cdot \vec{1} \right)\mathbf{T}_\lambda  + \mathbf{b}_\lambda^{(\text{LN})}
\end{align*}}
We can then consider what happens to this additive structure in the next sublayer.
We first define $\mathbf{T}_{\lambda+1}$ as previously and remark that, as both $\mathbf{T}_\lambda$ and $\mathbf{T}_{\lambda+1}$ only contain diagonal entries:
{\smaller\begin{equation*}
    \mathbf{T}_\lambda \mathbf{T}_{\lambda+1}= \frac{1}{s_{\lambda,t} s_{\lambda+1,t}} \begin{bmatrix}
        \left(\mathbf{g}_\lambda\odot\mathbf{g}_{\lambda+1}\right)_1 & \hdots & 0 \\
        \vdots & \ddots \\
        0 & & \left(\mathbf{g}_\lambda\odot\mathbf{g}_{\lambda+1}\right)_d \\
    \end{bmatrix}
\end{equation*}}
This generalizes for any sequence of LNs as:
{\smaller\begin{equation*}
    \prod\limits_\lambda \mathbf{T}_\lambda = \frac{1}{\prod\limits_\lambda s_{\lambda,t}} \begin{bmatrix}
        \left(\bigodot\limits_\lambda\mathbf{g}_\lambda\right)_1 & \hdots & 0 \\
        \vdots & \ddots \\
        0 & & \left(\bigodot\limits_\lambda\mathbf{g}_\lambda\right)_d \\
    \end{bmatrix}
\end{equation*}}

Let us now pass the input $\mathbf{x}$ through a complete layer, i.e., through sublayers $\lambda$ and $\lambda+1$:
{\smaller\begin{equation*}
    \mathbf{y}_{\lambda+1,t} =  S_{\lambda+1}\left(\mathbf{y}_{\lambda,t}\right)\mathbf{T}_{\lambda+1} + \mathbf{y}_\lambda\mathbf{T}_{\lambda+1} -  \left(m_{\lambda+1,t} \cdot \vec{1}\right) \mathbf{T}_{\lambda+1} + \mathbf{b}_{\lambda+1}^{(\text{LN})} 
\end{equation*}}
Substituting in the expression for $\mathbf{y}_{\lambda}$ from above: 
{\smaller\begin{align*}
    \mathbf{y}_{\lambda+1,t} =& S_{\lambda+1}\left( S_\lambda\left(\mathbf{x}\right)\mathbf{T}_\lambda + \mathbf{x}\mathbf{T}_\lambda - \left(m_{\lambda,t} \cdot \vec{1}\right)\mathbf{T}_\lambda   + \mathbf{b}_\lambda^{(\text{LN})}\right) \mathbf{T}_{\lambda+1} \\ 
    & +  S_\lambda\left(\mathbf{x}\right)\left(\prod \limits_{\lambda'=\lambda}^{\lambda+1} \mathbf{T}_{\lambda'}\right) +\mathbf{x} \left(\prod \limits_{\lambda'=\lambda}^{\lambda+1} \mathbf{T}_{\lambda'}\right) -  \left(m_{\lambda,t} \cdot \vec{1}\right)  \left(\prod \limits_{\lambda'=\lambda}^{\lambda+1} \mathbf{T}_{\lambda'}\right) \\
    & + \mathbf{b}_\lambda^{(\text{LN})}\mathbf{T}_{\lambda+1}  -  \left(m_{\lambda+1,t} \cdot \vec{1}\right)\mathbf{T}_{\lambda+1}  + \mathbf{b}_{\lambda+1}^{(\text{LN})} 
\end{align*}}

As we are interested in the combined effects of a layer, we only consider the case where $S_{\lambda}$ is a MHA mechanism and $S^{\lambda+1}$ a FF.
%
%
We start by reformulating the output of a MHA.
Recall that attention heads can be seen as weighted sums of value vectors \citep{kobayashi-etal-2020-attention}.
Due to the softmax normalization, attention weights $\alpha_{t,1}, \dots \alpha_{t,n}$ sum to 1 for any position $t$.
Hence: 
{\smaller\begin{align*}
    (\mathbf{A}_{h})_{t,\cdot} &= \sum_{t'=1}^n \alpha_{h,t,t'} (\mathbf{V}_{h})_{t',\cdot} = \sum_{t'=1}^n  \left[\alpha_{h,t,t'}  \mathbf{x}_{t'} \mathbf{W}_{h}^{(\text{V})} + \alpha_{h,t,t'} \mathbf{b}_{h}^{(\text{V})} \right] \\
    &= \left(\sum_{t'}^n \alpha_{h,t,t'}  \mathbf{x}_{t'}\mathbf{W}_{h}^{(\text{V})} \right) + \mathbf{b}_{h}^{(\text{V})} 
\end{align*}}
To account for all $H$ heads in a MHA, we concatenate these head-specific sums and pass them through the output projection $\mathbf{W}^{(\text{MHA,O})}$.
As such, we can denote the unbiased output of the MHA and the associated bias as: 
{\smaller
\begin{align*}
 \Tilde{\mathbf{h}}_{l,t} &= \sum_{h} \sum_{t'} \alpha_{l,h,t,t'} \mathbf{x}_{l,t'} \mathbf{Z}_{l,h} \\
 \mathbf{b}_l^{(\text{MHA})} &= \mathbf{b}_l^{(\text{MHA,O})} + \left( \bigoplus\limits_{h}^H \mathbf{b}_{l,h}^{(\text{V})}\right)\mathbf{W}_l^{(\text{MHA,O})}
\end{align*}}
with $\mathbf{Z}_{l,h}$ as introduced in \eqref{eq:h-term}.
By substituting the actual sublayer functions in our previous equation:
{\smaller\begin{align*}
    \mathbf{y}_{l,t} =& \Tilde{\mathbf{f}}_{l,t}\mathbf{T}_{\lambda+1} + \mathbf{b}_l^{(\text{FF,O})}\mathbf{T}_{\lambda+1} +  \Tilde{\mathbf{h}}_{l,t}\left(\prod \limits_{\lambda'=\lambda}^{\lambda+1} \mathbf{T}_{\lambda'}\right) \\
    &+ \mathbf{b}_l^{(\text{MHA})}\left(\prod \limits_{\lambda'=\lambda}^{\lambda+1} \mathbf{T}_{\lambda'}\right) +  \mathbf{x}\left(\prod \limits_{\lambda'=\lambda}^{\lambda+1} \mathbf{T}_{\lambda'}\right) -  \left(m_{\lambda,t} \cdot \vec{1}\right) \left(\prod \limits_{\lambda'=\lambda}^{\lambda+1} \mathbf{T}_{\lambda'}\right) \\
    &+ \mathbf{b}_\lambda^{(\text{LN})} \mathbf{T}_{\lambda+1} -  \left(m_{\lambda+1,t} \cdot \vec{1}\right)\mathbf{T}_{\lambda+1} + \mathbf{b}^{(\text{LN})}_{\lambda+1}
\end{align*}}
Here, given that there is only one FF for this layer, the output of sublayer function at $\lambda+1$ will correspond to the output of the FF for layer $l$, i.e., $\Tilde{\mathbf{f}}_{l,t} + \mathbf{b}_l^{(\text{FF,O})}$, and similarly the output for sublayer $\lambda$ should be that of the MHA of layer $l$, or $\Tilde{\mathbf{h}}_{l,t} + \mathbf{b}_l^{(\text{MHA})}$.
To match \cref{eq:global}, rewrite as:
{\smaller\begin{align*}
    \mathbf{y}_{l,t} =& ~ {\mathbf{i}}_{\lambda+1,t} + {\mathbf{h}}_{\lambda+1,t} + {\mathbf{f}}_{\lambda+1,t} + {\mathbf{c}}_{\lambda+1,t} \\
     {\mathbf{i}}_{\lambda+1,t} =&  \mathbf{x}_{\lambda,t} \left( \prod \limits_{\lambda'=\lambda}^{\lambda+1} \mathbf{T}_{\lambda'} \right)\\
     {\mathbf{h}}_{\lambda+1,t} =&  \Tilde{\mathbf{h}}_{l,t} \left( \prod \limits_{\lambda'=\lambda}^{\lambda+1} \mathbf{T}_{\lambda'} \right) \\
     {\mathbf{f}}_{\lambda+1,t} =& ~ \Tilde{\mathbf{f}}_{l,t} \mathbf{T}_{\lambda+1} 
\\
    {\mathbf{c}}_{\lambda+1,t} =& ~ \mathbf{b}_l^{(\text{FF,O})} \mathbf{T}_{\lambda+1} +  \mathbf{b}_l^{(\text{MHA})} \left( \prod \limits_{\lambda'=\lambda}^{\lambda+1} \mathbf{T}_{\lambda'} \right) \\
    &~ -  \left(m_{\lambda,t} \cdot \vec{1}\right) \left( \prod \limits_{\lambda'=\lambda}^{\lambda+1} \mathbf{T}_{\lambda'} \right) -  \left(m_{\lambda+1,t} \cdot \vec{1}\right) \mathbf{T}_{\lambda+1} \\ %
    &~+  \mathbf{b}_\lambda^{(\text{LN})} \mathbf{T}_{\lambda+1} + \mathbf{b}^{(\text{LN})}_{\lambda+1} %
\end{align*}}%
where $\mathbf{x}_{\lambda,t}$ is the $t$\textsuperscript{th} input for sublayer $\lambda$; i.e., the above characterizes the output of sublayer $\lambda+1$ with respect to the input of sublayer $\lambda$.
Passing the output $\mathbf{y}_{l,t}$ into the next layer $l+1$ (i.e., through sublayers $\lambda+2$ and $\lambda+3$) then gives:
{\smaller\begin{align*}
    \mathbf{y}_{l+1,t} =& {\mathbf{i}}_{\lambda+3,t} + {\mathbf{h}}_{\lambda+3,t} + {\mathbf{f}}_{\lambda+3,t} + {\mathbf{c}}_{\lambda+3,t} \\
    {\mathbf{i}}_{\lambda+3,t} =&  {\mathbf{i}}_{\lambda+1,t} \left(\prod \limits_{\lambda'=\lambda+2}^{\lambda+3} \mathbf{T}_{\lambda'}\right) \\
    {\mathbf{h}}_{\lambda+3,t} =&   {\mathbf{h}}_{\lambda+1,t} \left(\prod \limits_{\lambda'=\lambda+2}^{\lambda+3} \mathbf{T}_{\lambda'}\right) +   \Tilde{\mathbf{h}}_{l+1,t} \left(\prod \limits_{\lambda'=\lambda+2}^{\lambda+3} \mathbf{T}_{\lambda'}\right) \\
    {\mathbf{f}}_{\lambda+3,t} =&   {\mathbf{f}}_{\lambda+1,t} \left(\prod \limits_{\lambda'=\lambda+2}^{\lambda+3} \mathbf{T}_{\lambda'}\right) +  \Tilde{\mathbf{f}}_{l+1,t} \mathbf{T}_{\lambda+3}
\\
    {\mathbf{c}}_{\lambda+3,t} =&   {\mathbf{c}}_{\lambda+1,t} \left(\prod \limits_{\lambda'=\lambda+2}^{\lambda+3} \mathbf{T}_{\lambda'}\right) +   \mathbf{b}_l^{(\text{MHA})} \left(\prod \limits_{\lambda'=\lambda+2}^{\lambda+3} \mathbf{T}_{\lambda'}\right) + \mathbf{b}_l^{(\text{FF,O})} \mathbf{T}_{\lambda+3} \\
    &-  \left(m_{\lambda+2,t} \cdot \vec{1}\right) \left(\prod \limits_{\lambda'=\lambda+2}^{\lambda+3} \mathbf{T}_{\lambda'}\right) - \left(m_{\lambda+3,t} \cdot \vec{1}\right) \mathbf{T}_{\lambda+3}  \\
    &+ \mathbf{b}^{(\text{LN})}_{\lambda+2} \mathbf{T}_{\lambda+3}  + \mathbf{b}^{(\text{LN})}_{\lambda+3} 
\end{align*}}

This logic carries on across layers: adding a layer corresponds to (i) mapping the existing terms through the two new LNs, (ii) adding new terms for the MHA and the FF, (iii) tallying up biases introduced in the current layer.
Hence, the above generalizes to any number of layers $k \ge 1$ as:\footnote{
    The edge case $\prod_{\lambda'=\lambda+1}^{\lambda} \mathbf{T}_{\lambda'}$ is taken to be the identity matrix $\mathbf{I}_d$, for notation simplicity.
}
{\smaller\begin{align*}
    \mathbf{y}_{l+k,t} =& {\mathbf{i}}_{\lambda+2k-1,t} + {\mathbf{h}}_{\lambda+2k-1,t} + {\mathbf{f}}_{\lambda+2k-1,t} + {\mathbf{c}}_{\lambda+2k-1,t} \\
    {\mathbf{i}}_{\lambda+2k-1,t} =&  \mathbf{x}_{\lambda,t} \left(\prod \limits_{\lambda'=\lambda}^{\lambda+2k-1} \mathbf{T}_{\lambda'}\right) \\  
    {\mathbf{h}}_{\lambda+2k-1,t} =&  \sum \limits_{l'=l}^{l+k} \Tilde{\mathbf{h}}_{l',t} \left(\prod\limits_{\lambda'=2l'-1}^{2(l+k)} \mathbf{T}_{\lambda'}\right) \\
    {\mathbf{f}}_{\lambda+2k-1,t} =& \sum \limits_{l'=l}^{l+k} \Tilde{\mathbf{f}}_{l',t} \left(\prod\limits_{\lambda'=2l'}^{2(l+k)} \mathbf{T}_{\lambda'}\right)  
\end{align*}}
{\smaller\begin{align*}
    \!\!\!\!\!\!\!\!\!\!\!\!\!\!\!{\mathbf{c}}_{\lambda+2k-1,t} =&  \sum \limits_{\lambda'=\lambda}^{\lambda+2k-1} \left[  \mathbf{b}^{(\text{LN})}_{\lambda'} \left( \prod_{\lambda''=\lambda'+1}^{\lambda+2k-1} \mathbf{T}_{\lambda''}\right) -  \left(m_{\lambda',t} \cdot \vec{1}\right) \left(\prod_{\lambda''=\lambda'}^{\lambda+2k-1} \mathbf{T}_{\lambda''}\right) \right]  \\
                                    &+ \sum \limits_{l'=l}^{l+k} \left[  \mathbf{b}_{l'}^{(\text{MHA})} \left(\prod_{\lambda'=2l-1}^{\lambda+2k-1} \mathbf{T}_{\lambda'}\right) +  \mathbf{b}_{l'}^{(\text{FF,O})} \left(\prod_{\lambda'=2l-1}^{\lambda+2k-1} \mathbf{T}_{\lambda'}\right) \right]
\end{align*}}

Lastly, recall that by construction, we have:
{\smaller\begin{equation*}
     \mathbf{v} \left(\prod\limits_\lambda \mathbf{T}_\lambda \right)= \frac{\bigodot\limits_\lambda \mathbf{g}_\lambda }{\prod\limits_\lambda s_{\lambda,t}} \odot \mathbf{v}
\end{equation*}}
By recurrence over all layers and providing the initial input $\mathbf{x}_{0,t}$, we obtain \cref{eq:global,eq:i-term,eq:h-term,eq:f-term,eq:c-term}.

\section{Hyperplane bounds of $\mathbf{c}_t$}
\label{sec:appendix-hyperplane}
We can re-write \cref{eq:c-term} to highlight that is comprised only of scalar multiplications applied to constant vectors.
Let:
{\smaller\begin{align*}
    \mathbf{b}_\lambda^{(\text{S})} &= 
    \begin{cases}
        \mathbf{b}_l^{(\text{MHA,O})} + \left(\bigoplus\limits_{h} \mathbf{b}_{l,h}^{(\text{V})}\right) \mathbf{W}_l^{(\text{MHA,O})} & ~ \textit{if} ~ \lambda=2l-1 \\
        \mathbf{b}_l^{(\text{FF,O})} & ~ \textit{if} ~ \lambda=2l
    \end{cases} \\
    \mathbf{p}_\lambda &= \left(\bigodot\limits_{\lambda'=\lambda+1}^{\Lambda} \mathbf{g}_{\lambda'} \right) \odot (\mathbf{b}_\lambda^{(\text{LN})} + \mathbf{b}_\lambda^{(\text{S})}) \\
    \mathbf{q}_\lambda &= \bigodot\limits_{\lambda'=\lambda+1}^{\Lambda} \mathbf{g}_{\lambda'} 
\end{align*}}
Using the above, \cref{eq:c-term} is equivalent to:
{\smaller\begin{align*}
	\mathbf{c}_t = \sum_\lambda^\Lambda \left( \frac{1}{\prod\limits_{\lambda'=\lambda+1}^{\Lambda} s_t^{\lambda'}}  \cdot \mathbf{p}_\lambda \right) + \sum_\lambda^\Lambda \left( \frac{-m_{\lambda,t}}{\prod\limits_{\lambda'=\lambda+1}^{\Lambda} s_{\lambda',t}} \cdot \mathbf{q}_\lambda \right)
\end{align*}}
Note that $\mathbf{p}_\lambda$ and $\mathbf{q}_\lambda$ are constant across all inputs. 
Assuming their linear independence puts an upper bound of $2\Lambda$ vectors necessary to express $\mathbf{c}_t$.

\section{Computational details}

In \Cref{subsec:limits}, we use the default hyperparameters of \texttt{scikit-learn} \citep{scikit-learn}. 
In \Cref{sec:exp-mlm}, we learn categorical regressions using an AdamW optimizer \citep{loshchilov-hutter-2018-decoupled} and iterate 20 times over the train set; hyperparameters (learning rate, weight decay, dropout, and the $\beta_1$ and $\beta_2$ AdamW hyperparameters) are set using Bayes Optimization \citep{snoek-etal-2012-practical}, with 50 hyperparameter samples and accuracy as objective.
In \Cref{sec:exp-wsd}, learning rate, dropout, weight decay, $\beta_1$ and $\beta_2$, learning rate scheduling are selected with Bayes Optimization, using 100 samples and accuracy as objective.
In \Cref{sec:exp-ner}, we learn shallow logistic regressions, setting hyperparameters with Bayes Optimization, using 100 samples and macro-$f_1$ as the objective.
Experiments were run on a 4GB NVIDIA GPU. 

\section{Ethical considerations}
The offset method of \citet{mikolov-etal-2013-linguistic} is known to also model social stereotypes \citep[a.o.]{NIPS2016_a486cd07}.
Some of the sub-representations of our decomposition may exhibit stronger biases than the whole embedding $\mathbf{e}_t$, and can yield higher performances than focusing on the whole embedding (e.g., \Cref{tab:wsd-acc}).
This could provide an undesirable incentive to deploy NLP models with higher performances and stronger systemic biases. 

\end{document}